\documentclass{article}

\usepackage{arxiv}
\usepackage{natbib}
\usepackage{amsmath,amssymb}
\usepackage{booktabs}
\usepackage{graphicx}
\usepackage{wrapfig}
\usepackage{microtype}
\usepackage{multirow}
\usepackage{url}
\usepackage{hyperref}
\title{Pepti-drift: Scalable Safe--Active Peptide Generation Without Inference-Time Guidance}
\author{Takashi Fujiwara\\
	    Department of AI Systems Medicine \\
        Institute of Biomedical Engineering\\
        Institute of Science Tokyo,\\
        SB Intuitions\\
	\And
	Hikaru Shindo\thanks{\texttt{\{hikaru.shindo, keisuke.ozawa\}@sbintuitions.co.jp}} \\
	SB Intuitions \\
    \And
	Kaushalya Madhawa \\
	SB Intuitions\\
    \And
	Jun Jin Choong \\
	SB Intuitions\\
    \And
    Shuan Chen \\
	SB Intuitions\\
    \And
    Yuna Oikawa \\
    The University of Tokyo,\\
	SB Intuitions\\
    \And
    Yiming Zhang \\
    The University of Tokyo,\\
	SB Intuitions\\
    \And
    Gyubok Lee \\
	SB Intuitions\\
    \And
	Keisuke Ozawa$^*$ \\
	SB Intuitions
}
\renewcommand{\undertitle}{}   
\renewcommand{\headeright}{}   
\newcommand{\benchmark}{\textsc{BindSafe-PepBench}}
\newcommand{\safeactive}{\textsc{Safe--Active}}
\newcommand{\bindscore}{s_{\mathrm{bind}}}

\begin{document}
\maketitle

\begin{abstract}
    Therapeutic peptides are a promising drug modality, but their generation must satisfy multiple therapeutic constraints.
    We introduce \textbf{BindSafe-PepBench}, a fixed-budget benchmark that jointly evaluates target binding and four major safety metrics on the same generated candidates.
    We reveal that peptide length is a major confounder of joint binding–safety evaluation: longer peptides tend toward stronger predicted binding but less favorable predicted safety. This creates an apparent trade-off and can bias comparisons among models with different output-length distributions. We report absolute \safeactive{} yield and exact-length-matched gains to distinguish generative improvements from output-length effects.
    High \safeactive{} yield remains challenging, while the strongest multi-property methods rely on costly inference-time guidance.
    We therefore introduce \textbf{Pepti-drift}, a one-step generation framework that incorporates attraction toward target-specific binders and repulsion from liability-associated regions, requiring a single latent refinement followed by parallel decoding without inference-time guidance.
    Across 88 held-out targets, Pepti-drift achieves an 18.37\% predicted \safeactive{} yield while retaining positive exact-length-matched gains. The resulting gains are competitive with multi-property-guided baselines while requiring \textbf{468$\times$ lower generation cost}, enabling scalable and fair high-throughput peptide design.
\end{abstract}

\section{Introduction}

\begin{figure*}[t]
\centering
\includegraphics[width=0.97\linewidth]{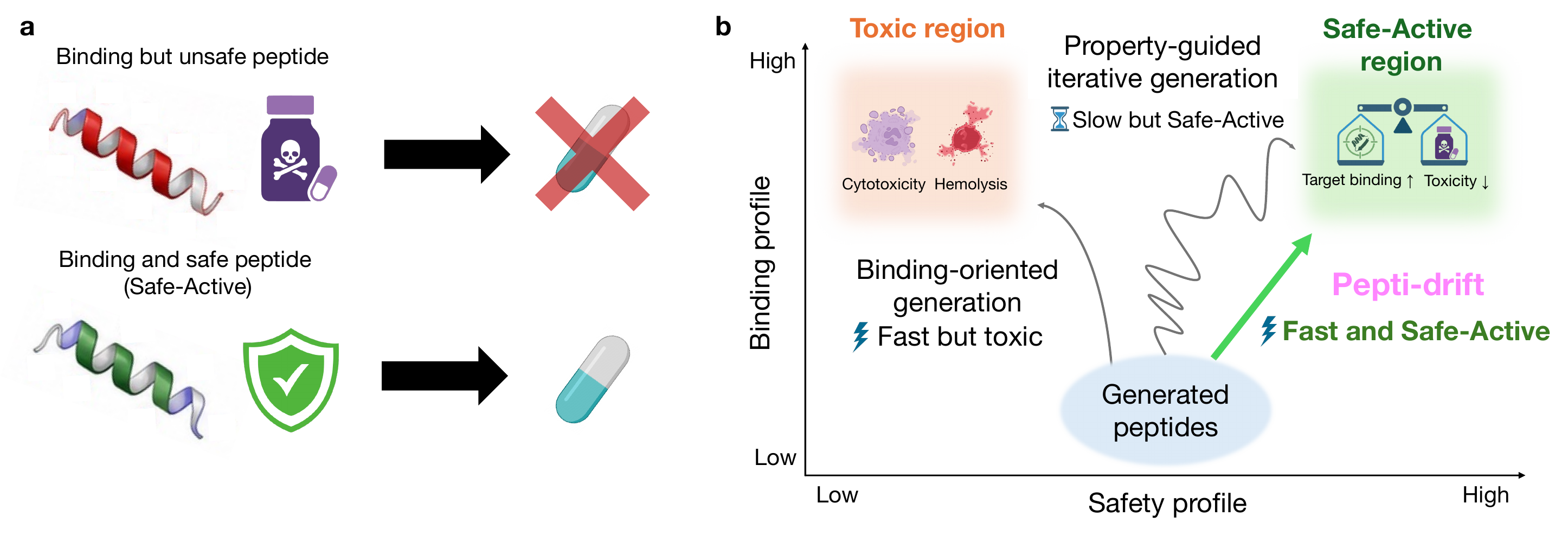}
\vspace{-1em}
\caption{
\textbf{Binding alone is insufficient for therapeutic peptide generation.}
(a) A therapeutic candidate must satisfy both binding and safety.
(b) Binding-oriented generation may overlook safety, whereas iterative
property-guided methods improve joint control at additional inference cost. Pepti-drift instead amortizes multi-property control into training.
}
\vspace{-1em}
\label{fig:drift_intro}
\end{figure*}

Peptide therapeutics act by binding to specific sites on disease-relevant proteins and modulating their function \citep{muttenthaler2021trends,wang2022therapeutic}.
Recent advances in generative modeling provide a powerful way of exploring the vast peptide space and proposing target-specific binders \citep{chen2026pepmlm,kong2024pepglad}.
However, therapeutic peptide generation is not merely a binder search problem; viable candidates must also avoid major safety liabilities, including toxicity, hemolysis, allergenicity, and neurotoxicity \citep{wang2022therapeutic,rathore2025prediction}.
Safety is central to drug development and has been a major source of clinical attrition \citep{Kola2004,Arrowsmith2013,kuo2024consideration,filipova2026need}.
Strong binding and favorable safety must be achieved simultaneously (Figure~\ref{fig:drift_intro}a): a strong binder that causes hemolysis is not a viable candidate, just as a safe but nonbinding peptide is ineffective.
Yet, existing evaluation protocols lack such joint assessment: binding or structural quality is typically evaluated independently of safety properties \citep{kong2024pepglad,kong2024pepbenchdata,zhang2026pepbenchmark,zhou2026ampbench}. 
This fragmentation makes it difficult to evaluate progress toward therapeutically viable peptide design consistently and fairly.

To address this gap, we introduce \textbf{BindSafe-PepBench}, a fixed-budget benchmark that jointly evaluates binding and four major safety liabilities on the same generated candidates.
Applying this benchmark reveals an apparent negative binding--safety association that is largely explained by peptide length: longer peptides tend to exhibit stronger predicted binding but less favorable predicted safety.
This finding motivates reporting both absolute \safeactive{} yield and exact-length-matched enrichment to distinguish genuine generative improvements from output-length effects.

Achieving high \safeactive{} yield remains challenging.
Recent property-aware methods such as PepTune \citep{tang2025peptune} and MOG-DFM \citep{chen2025mogdfm} steer generation toward multiple therapeutic objectives
through iterative inference-time guidance. More generally, guided diffusion and flow methods have demonstrated the effectiveness of optimizing objectives during the generative trajectory, including in peptide design \citep{wang2025ocflow}. 
While effective, these methods repeatedly query guidance signals during sampling, incurring substantial computational cost. This overhead limits the number of candidates, targets, and design iterations that can be explored. This motivates our central research question: \emph{How can we efficiently generate peptides that satisfy both binding and safety without costly inference-time optimization?}


To answer this question, we introduce \textbf{Pepti-drift}, a framework that amortizes multi-property control into training rather than repeatedly enforcing it during inference.
Building on Drifting Models~\citep{deng2026drifting}, Pepti-drift encodes attraction toward target-specific binders and repulsion from safety liabilities in a learned latent-space vector field, avoiding repeated optimization during sampling (Figure~\ref{fig:drift_intro}b).
Given only a target protein sequence and sampled noise, Pepti-drift performs a single latent refinement followed by parallel decoding. Generation requires neither target structures nor iterative queries to binding or safety property models.


This design shifts the joint quality--cost frontier. Across 88 held-out targets, Pepti-drift achieves an 18.37\% \safeactive{} yield, nearly matching the highest observed yield of 18.45\%. It requires only 2.643 ms per peptide, corresponding to a $468\times$ lower generation cost than the strongest inference-guided multi-property baseline.
It also retains positive exact-length-matched enrichment, indicating that its performance cannot be explained by output-length differences alone. Qualitative structural analysis further suggests that
successful Pepti-drift outputs form well-buried target interfaces rather than exploiting simple one-dimensional compositional shortcuts.

These results show that joint control can be learned during training without costly inference-time search, shifting the quality--cost frontier toward faster and more practical peptide generation.

Our key contributions are threefold:
\begin{enumerate}
\item \textbf{Joint evaluation of binding and safety.}
We introduce \textsc{BindSafe-PepBench}, a fixed-budget benchmark that evaluates joint target binding and four major safety constraints on the same generated peptide.

\item \textbf{Identification of peptide length as a major confounder.}
We show that peptide length largely explains the apparent binding--safety trade-off, motivating exact-length-matched evaluation to separate generative improvements from length effects.

\item \textbf{Efficient joint control without inference-time optimization.}
We introduce \textsc{Pepti-drift}, which amortizes multi-property control into training and achieves near-best raw joint yield at 468$\times$ lower generation cost than the strongest inference-guided multi-property baseline.

\end{enumerate}

\section{Related Work}
\label{sec:related_work}

\paragraph{Generative modeling for biomolecular design.}
Generative models enable biomolecule design with desired structures and properties.
Representative approaches include
protein language models for sequence generation
\citep{nijkamp2023progen2,wang2024dplm,hayes2025esm3},
diffusion models for structure generation
\citep{yim2023framediff,watson2023rfdiffusion},
and goal-directed molecular optimization
\citep{you2018gcpn,bengio2021gflownet,jain2023mogfn}.
Our work focuses on efficient multi-property control for target-conditioned peptide generation.


\paragraph{Target-conditioned therapeutic peptide design.}
Structure-aware peptide generators such as PepGLAD
\citep{kong2024pepglad}, PepFlow \citep{li2024pepflow},
PPFlow \citep{lin2024ppflow}, DiffPepBuilder
\citep{wang2024diffpepbuilder}, PepHAR \citep{li2025pephar}, and PepTri \citep{nguyen2026peptri} condition on target structures, binding sites, or related geometric information.
PepMimic \citep{kong2025peptide} additionally conditions peptide sequence--structure co-design on the binding interface of a protein binder.
Beyond peptides, BindCraft \citep{pacesa2024bindcraft, pacesa2025bindcraft} generates protein binders through iterative structure–sequence co-optimization.
In contrast, PepMLM \citep{chen2026pepmlm} and High-PepBinder
\citep{mao2026highpepbinder} generate target-specific peptides directly
from protein sequences.
Multi-property methods such as PepTune \citep{tang2025peptune} and
MOG-DFM \citep{chen2025mogdfm} further steer generation through inference-time property guidance, requiring repeated property evaluations during sampling.
Related approaches to reducing iterative sampling cost include progressive distillation, consistency models, and rectified flow \citep{salimans2022progressive,song2023consistency,liu2023rectified,yin2024one}.
Drifting Models \citep{deng2026drifting} have also been applied to molecular generation and conformation sampling \citep{qiu2026driftingmol,hu2026drifting}.
Pepti-drift extends this framework to target-conditioned peptide generation, incorporating attraction toward target-specific binders and repulsion from liability-associated regions during training to avoid iterative property queries at inference.

\paragraph{Benchmarks for peptide design.}
PepBench \citep{kong2024pepbenchdata} provides datasets and splits for
protein-binding peptide design, PPBench2024 supports structure-based peptide
design \citep{lin2024ppflow}, and PepBenchmark
\citep{zhang2026pepbenchmark} standardizes predictive evaluation across
peptide properties.
Broader biomolecular and molecular-design benchmarks emphasize
multidimensional evaluation, reproducibility, oracle efficiency, or
computational throughput
\citep{ye2025proteinbench,lin2025cbgbench,brown2019guacamol,
gao2022pmo,liu2026protdbench}.
However, these protocols do not directly assess whether the same generated
peptide jointly satisfies target binding and multiple safety constraints under a fixed candidate budget.
\benchmark{} addresses this gap by jointly reporting \safeactive{} yield, exact-length-matched enrichment, and generation cost.

\section{\benchmark{}}
\label{sec:evaluation}

A useful therapeutic peptide must satisfy two requirements simultaneously:
it should bind the intended protein target and avoid major safety liabilities.
However, existing evaluations often assess binding and safety separately,
leaving unclear whether the \emph{same generated peptide} satisfies both.
We therefore introduce \benchmark{}, a fixed-budget benchmark that evaluates
joint binding and safety on individual generated candidates under a common
evaluation protocol.

\textbf{Fixed-budget joint generation and evaluation.}
Experimental validation is resource-intensive, and precise 3D target structures are often unavailable. \benchmark{} therefore considers sequence-conditioned \emph{de novo} generation: given only a target protein sequence, each method generates a fixed set of peptides without a predefined binding site or target structure.
We allocate $K=30$ candidate slots per target, a moderate fixed budget within the range used in recent peptide-generation evaluations
\citep{lin2024ppflow,huang2026epic}.
We restrict candidates to 4--25 amino acids, following the peptide-length rule adopted in PepBench \citep{kong2024pepbenchdata}, rather than introducing a model-specific length range.
The benchmark then asks whether each candidate passes a single joint endpoint, \safeactive{}.
All methods are rescored using the same external evaluators rather than their own training rewards or guidance models, avoiding direct preference for any generator's internal objectives.

For predicted binding, we first use ESMFold2
\citep{synthyra2026esmfold2} to predict the 3D target--peptide complex.
To avoid interpreting binding scores from low-confidence predicted complexes,
we require $\mathrm{ipTM}\geq0.5$, reflecting confidence in the relative
target--peptide arrangement, and mean interface $\mathrm{pLDDT}\geq70$,
reflecting local structural confidence around the predicted interface.
PRODIGY \citep{vangone2015contact,xue2016prodigy} provides
$\bindscore=-\Delta G$.
For candidate-level evaluation, scores are retained only after
confidence gating, and activity requires $\bindscore\geq7$.

For predicted safety, four ESM-2-based classifiers independently evaluate toxicity, hemolysis, allergenicity, and neurotoxicity.
A candidate must have a predicted liability score of at most 0.5 for all four properties.
Finally, it must satisfy basic sequence-integrity requirements, including canonical amino acids, the allowed length range, within-target uniqueness, and exact novelty relative to the generator's training peptides.
A peptide is counted as \safeactive{} only when all binding, safety, and
integrity criteria are satisfied simultaneously.
These outcomes are computational predictions rather than experimental measurements; complete endpoint definitions and evaluator performance are reported in Appendix~\ref{app:evaluators}.





\textbf{Joint yield and length-matched enrichment.}
Our primary metric is raw \safeactive{} yield, defined as the fraction of the fixed $K=30$ candidate slots that satisfy the complete joint endpoint.
Because every allocated slot remains in the denominator, failed or unusable generations directly reduce this yield.
Raw yield can also depend on a generator's output-length distribution, because length affects both predicted binding and safety.
We therefore report \emph{Exact-length-matched enrichment@$K$} alongside raw yield.
For each submitted candidate, we construct a random control peptide matched to
the same target, submitted rank, and exact peptide length. We then compute the
difference in \safeactive{} rate between the generated candidates and their
matched random controls for each target and average this difference across
targets. Enrichment is reported in percentage points; a positive value means
that the generator produces more \safeactive{} candidates than random
sequences under the same target and exact-length schedule.
We additionally report \safeactive{} target coverage, defined as the fraction
of targets for which at least one \safeactive{} candidate is found, together
with marginal binding performance and generation cost. Further metric and
control details are provided in Appendix~\ref{app:metrics}.

\section{Pepti-drift}
\label{sec:peptidrift}

As motivated by \benchmark{}, therapeutic peptide generation requires joint control of binding and safety under a fixed candidate budget.
Pepti-drift amortizes this multi-property control into training using the drifting paradigm~\citep{deng2026drifting}. It learns a vector field that moves target-conditioned latent representations toward target-specific binders and away
from liability-associated regions. At inference, generation requires only a
single latent refinement followed by parallel non-autoregressive decoding,
without iterative property evaluation or search.


\begin{figure}[t]
    \centering
    \includegraphics[width=0.98\textwidth]
    {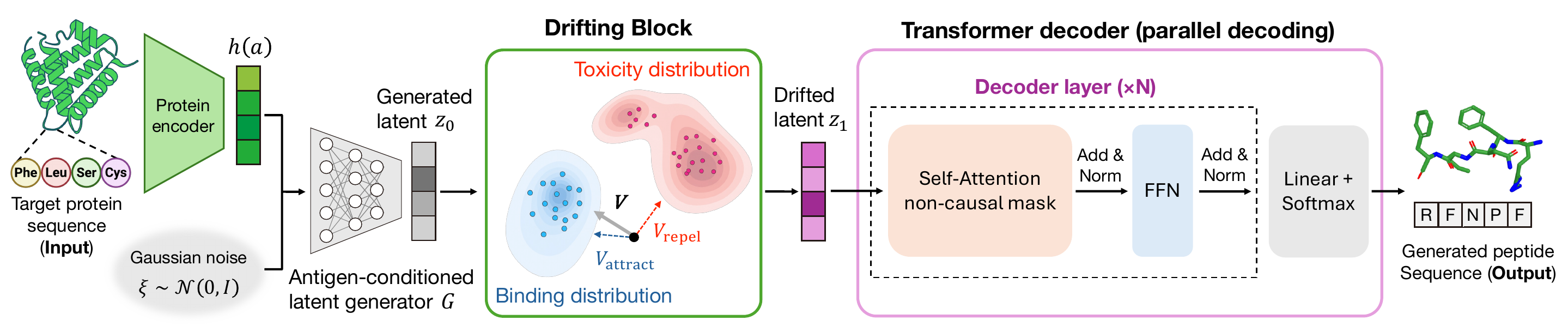}
    \vspace{-1em}
    \caption{
    \textbf{Overview of the Pepti-drift framework.} A target sequence $a$ is encoded by a protein language model, and a target-conditioned generator maps its representation and Gaussian noise to an initial peptide latent $z_0$. A drifting block applies a learned vector
    field $V=V_{\mathrm{attract}}+V_{\mathrm{repel}}$, combining attraction toward the target-specific binding-peptide distribution with repulsion from liability-associated peptide regions, to obtain the refined latent $z_1$.
    A non-autoregressive Transformer decoder then generates the peptide sequence from $z_1$ in a single parallel pass.}
    \label{fig:overview}
\end{figure}



\textbf{One-step target-conditioned generation.}
Pepti-drift maps the target protein representation and sampled noise to an initial latent, followed by a single refinement step (Figure~\ref{fig:overview}).

Let $a$ denote a target protein sequence and $y$ a reference peptide sequence. 
A frozen protein language model (ESM-2) produces the base target and peptide embeddings \citep{lin2023esm}. The target embedding is passed through a trainable projection $h(a)$, whereas peptide references are mapped to normalized latent vectors $u_0(y) \in \mathbb{S}^{d_z-1}$ using a fixed, randomly initialized projection, where $d_z=1024$. This fixed reference map is used exclusively to construct training targets.

An initial generator $G$ maps the target representation and Gaussian noise $\xi\sim\mathcal{N}(0,I)$ to an initial latent $z_0$.
A trainable drift module $D$ then applies a single learned update, yielding the final drifted latent $x$ (equivalently $z_1$),
\begin{equation}
z_0 = G(h(a), \xi), \qquad x = z_1 = z_0 + \eta D(z_0, h(a)), \quad \eta=1.
\end{equation}
At inference, the drift module $D$ is evaluated only once during the forward pass.



\textbf{Attraction--repulsion training objective.}
To jointly control binding and safety, we construct a dynamic training target that combines attraction toward a reference binder with repulsion from liability-associated peptides.
We first \(\ell_2\)-normalize the current generated latent \(x\) as
\(\bar{x}=\operatorname{norm}(x)=x/\lVert x\rVert_2\), placing it in the same normalized latent space as the peptide references.
Instead of scalar rewards, we construct a spatial target
$\widetilde{x}$ for the current normalized generated latent $\bar{x}$ using
\begin{equation}
V(\bar{x})
=
\alpha_+ V_{\text{attract}}(\bar{x})
+
\alpha_- V_{\text{repel}}(\bar{x}).
\end{equation}
The \emph{attractive component} ($V_{\text{attract}}$) moves the representation toward a target-matched
reference binder, whereas the \emph{repulsive component} ($V_{\text{repel}}$) moves it away from nearby
toxicity- and hemolysis-associated references. The \safeactive{} endpoint evaluates four safety liabilities,
whereas Pepti-drift constructs its repulsive field only from toxicity- and hemolysis-associated training examples. Allergenicity and neurotoxicity are kept evaluation-only, providing independent safety axes for testing generalization beyond the liabilities explicitly used during training.


We dynamically mine hard negatives whose embeddings are closest to the current latent and weight their repulsive contributions by distance. The resulting training target is $\widetilde{x} = \operatorname{sg} \left[ \operatorname{norm} \left( \bar{x} + V(\bar{x}) \right) \right]$, where $\operatorname{sg}$ denotes stop-gradient.
The attraction coefficient is
fixed, while the repulsion strength is gradually increased during training.

A non-causal transformer decodes the refined latent and target representations into all peptide positions in parallel, with sequence length determined by the first predicted \texttt{<eos>} token. Training minimizes a joint objective balancing three weighted components: a \emph{sequence reconstruction loss} (token-level cross-entropy over non-padding positions), a \emph{drift-alignment loss} aligning the normalized generated latent with the spatial target, and a \emph{displacement regularizer} penalizing excessively large shifts during the one-step drift. The binding and safety evaluators used in \benchmark{} are never used as training rewards. At inference, generation requires only the target protein sequence and sampled noise. Full target construction, decoder, and optimization details are in Appendix~\ref{app:model}.

\section{Experiments}
\label{sec:experiments}

We investigate the relationship between predicted binding and safety under the fixed-budget \benchmark{} protocol and evaluate whether Pepti-drift can shift the quality--cost frontier of multi-property peptide generation through the following research questions:

\textbf{(Q1)} Is there a systematic tension between predicted binding and safety, and what factors drive it? \\
\textbf{(Q2)} Does Pepti-drift achieve joint success without sacrificing either the binding or safety axis? \\
\textbf{(Q3)} Can Pepti-drift match the joint \safeactive{} yield of state-of-the-art guided methods at a fraction of the computational cost? \\
\textbf{(Q4)} Does Pepti-drift find more \safeactive{} candidates per reached target than existing methods? \\
\textbf{(Q5)} Is Pepti-drift's joint success more consistent with superficial reward hacking or physically grounded 3D interactions?


We disentangle the factors driving apparent binding--safety trade-offs, revealing peptide length as a major source of evaluation bias, and demonstrate that Pepti-drift achieves a competitive \safeactive{} yield without relying on structural priors or expensive post-hoc optimization.

\subsection{Experimental setup}
\label{sec:experimental_setup}

\paragraph{Datasets and leakage control.}
The binding dataset contains 4,263 target--peptide complexes:
4,070 training, 105 validation, and 88 held-out test examples
\citep{kong2024pepbenchdata}.
Removing exact test-peptide matches leaves 4,060 training pairs.
The property dataset contains 25,480 PepBenchmark records
\citep{zhang2026pepbenchmark}.
Toxicity, hemolysis, allergenicity, and neurotoxicity define the four safety liabilities used in the \safeactive{} endpoint.
Pepti-drift's repulsion pool is constructed from toxicity- and hemolysis-labeled training peptides and is filtered against the property test set and all binding-positive peptides, leaving 3,577 sequences with no exact overlap with either protected set.
Full dataset construction and leakage safeguards are described in Appendix~\ref{app:data}.


\paragraph{Baselines.}
We compare Pepti-drift with target-conditioned generators spanning
different generation and control strategies. PepMLM~\citep{chen2026pepmlm} and
High-PepBinder~\citep{mao2026highpepbinder} generate target-specific peptides without inference-time guidance, whereas PepTune~\citep{tang2025peptune} and MOG-DFM~\citep{chen2025mogdfm} use
multi-property guidance during inference.
We additionally include a conditional autoregressive Transformer baseline trained on the same target representations and binding data as Pepti-drift. 
Because the primary benchmark is sequence-conditioned, methods requiring target structures or predefined binding pockets are treated as complementary rather than directly comparable baselines. Comparator configurations are provided in Appendix~\ref{app:reproducibility}.

To ensure fair evaluation, Pepti-drift and the trainable controls use benchmark-aligned splits with strict leakage safeguards. For published baselines using released checkpoints, however, exact separation between their original training data and \benchmark{} evaluation data cannot be verified.

\paragraph{Metrics.}
The primary metric is raw \safeactive{} yield at $K=30$, interpreted alongside exact-length-matched enrichment@$K$, which measures improvement over target- and length-matched random controls.
These metrics capture absolute joint success and improvement over a matched baseline, respectively.
We also report \safeactive{} target coverage, the marginal binding-active rate, and generation cost in milliseconds per peptide. \safeactive{} target coverage is the fraction of targets with at least one \safeactive{} candidate, while binding-active rate is the fraction of submitted candidate rows satisfying the structure-confidence criteria and the binding threshold. Additional marginal binding diagnostics are reported in Appendix \ref{app:fixedbudget}.

\subsection{Peptide length largely explains the apparent binding--safety trade-off}
\label{sec:tradeoff_main}

\begin{figure*}[t]
\centering
\includegraphics[width=0.9\linewidth]
{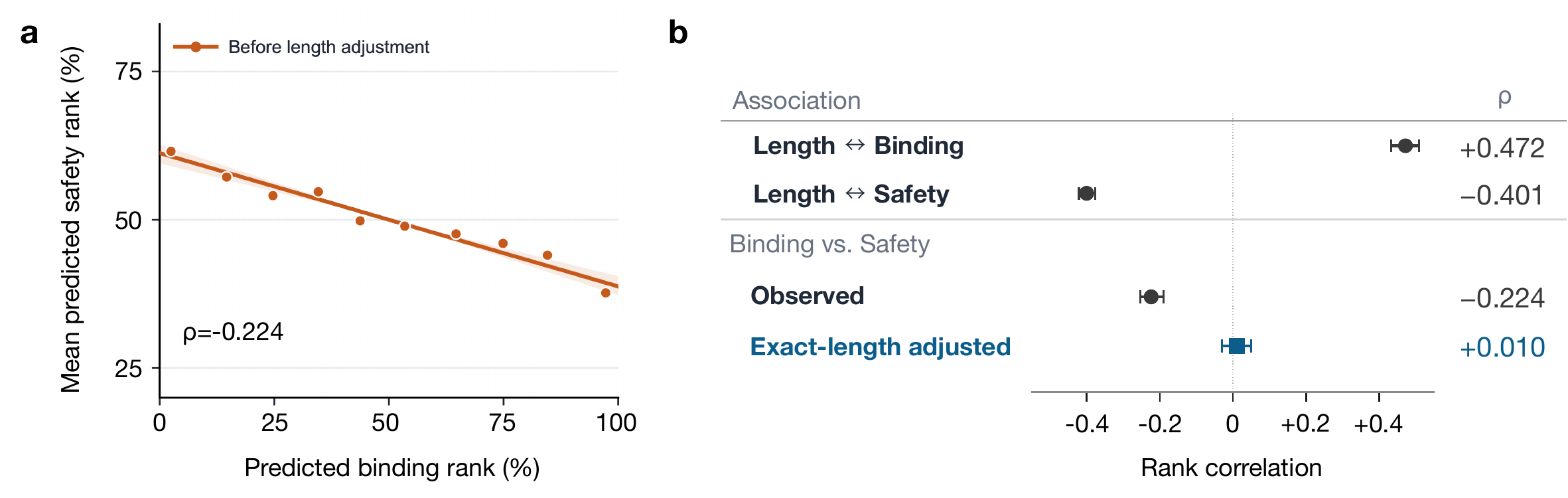}
\vspace{-1em}
\caption{
\textbf{Peptide length is a major driver of the apparent binding--safety tension.}
(a) Predicted binding and safety exhibit a negative association before accounting for peptide length ($\rho=-0.224$).
(b) Peptide length is positively associated with predicted binding and negatively associated with predicted safety. Accounting for exact peptide length nearly eliminates the observed association ($\rho=+0.010$). Error bars show target-bootstrap 95\% confidence intervals.
}
\label{fig:tradeoff}
\end{figure*}

To answer \textbf{(Q1)}, we examine whether predicted binding and safety exhibit a systematic tension, and what factors drive this relationship.
During preliminary analysis, we observed differences in peptide-length distributions across models, suggesting that length may confound both multi-property evaluation and model comparison.
To remove between-method and between-target score shifts, we rank predicted binding and the four-liability safety margin within each method--target stratum before pooling. Of 6,568 complete-case candidates, 6,356 belong to estimable strata and enter the analysis.

Before accounting for peptide length, stronger predicted binding is associated with less favorable predicted safety (Figure~\ref{fig:tradeoff}a; Spearman $\rho=-0.224$), indicating an apparent tension. 
Peptide length correlates oppositely with the two axes (Figure~\ref{fig:tradeoff}b): longer peptides tend to receive more favorable predicted binding scores ($\rho=+0.472$) but less favorable safety margins ($\rho=-0.401$).

This shared dependence largely explains the apparent binding--safety trade-off. When we  stratify candidates by exact peptide length, restricting the analysis to method--target--length strata with sufficient observations ($n=3{,}859$), the binding--safety association is nearly eliminated ($\rho=+0.010$). 
Thus, the raw negative correlation does not reflect an inherent, length-independent binding--safety antagonism.
Instead, peptide length emerges as a major driver of the relationship.

This finding directly motivates the length-controlled component of \benchmark{}. Raw \safeactive{} yield remains the primary measure of absolute joint success under the fixed candidate budget, but it can be affected by a generator's output length distribution. We therefore interpret it jointly with Exact-length-matched enrichment@$K$, which asks whether a generator outperforms random peptides under the same target and exact-length schedule.

Both the unadjusted negative association and its dependence on peptide length are robust across evaluators. Across all eight binding- or safety-oracle configurations, exact-length stratification reduces the correlation to near zero ($\rho=-0.014$ to $+0.032$; all 95\% CIs include zero; Appendix~\ref{app:oracle_length}).
The unadjusted correlation is stronger for more length-dependent evaluators.
These analyses answer \textbf{(Q1)}: the apparent negative association between predicted binding and safety is largely explained by peptide length rather than a length-independent antagonism, motivating exact-length-matched evaluation.

\subsection{Pepti-drift remains competitive on both marginal binding and safety axes}
\label{sec:marginal_axes}

To answer \textbf{(Q2)}, we examine whether Pepti-drift remains competitive on both the predicted binding and safety axes before applying the joint \safeactive{} criterion.
Table~\ref{tab:marginal_axes} summarizes the marginal quantities most directly aligned with the benchmark gates; complete safety and binding diagnostics are provided in Appendix~\ref{app:fixedbudget}.


\begin{table}[t]
\setlength{\belowcaptionskip}{4pt}
\caption{
\textbf{Pepti-drift remains competitive on both safety and binding axes.}  Marginal performance at $K=30$. Random uniform-length samples both peptide length and amino acids at random. Safety metrics report Safe-only and individual safety-gate pass rates. Binding metrics report binding coverage, binding-active rate, and target-level binding success. Bold and underlining mark the best and second-best among noncollapsed learned or guided methods.
}
\label{tab:marginal_axes}
\centering
\scriptsize
\setlength{\tabcolsep}{2.8pt}
\renewcommand{\arraystretch}{1.04}
\begin{tabular}{lrrrrrrrr}
\toprule
& \multicolumn{5}{c}{Predicted safety $\uparrow$}
& \multicolumn{3}{c}{Predicted binding $\uparrow$} \\
\cmidrule(lr){2-6}
\cmidrule(lr){7-9}
Method &
Safe-only &
Tox &
Hemo &
All. &
Neuro &
Bind cov. &
Bind-active &
Target pass \\
\midrule

\textbf{Pepti-drift (ours)}
& \underline{61.17}
& \underline{90.61}
& \underline{94.55}
& 80.64
& \underline{82.23}
& \textbf{35.45}
& \underline{25.80}
& \textbf{26.14} \\

PepTune
& \textbf{72.73}
& \textbf{96.86}
& \textbf{96.14}
& \textbf{88.75}
& \textbf{90.83}
& \underline{32.54}
& 24.32
& \underline{25.00} \\

MOG-DFM
& 57.90
& 88.76
& 90.02
& \underline{88.42}
& 82.04
& 23.46
& 18.75
& 15.91 \\

High-PepBinder
& 42.88
& 78.41
& 88.18
& 78.52
& 64.62
& 25.15
& 19.36
& 17.05 \\

PepMLM
& 34.09
& 72.20
& 88.41
& 69.39
& 62.77
& 31.10
& \textbf{27.16}
& \underline{25.00} \\

AR Transformer
& 51.93
& 84.13
& 91.48
& 78.18
& 75.49
& 29.13
& 23.37
& \underline{25.00} \\

Random uniform-length
& 35.68 
& 71.33 
& 70.11 
& 84.85 
& 57.39 
& 20.64
& 16.74
& 9.09
\\

\bottomrule
\end{tabular}
\end{table}

The principal generators occupy distinct marginal operating points
(Table~\ref{tab:marginal_axes}). PepTune is strongest on the safety axis,
achieving the highest Safe-only rate (72.73\%) and the highest pass rates for
all four individual safety gates. PepMLM, in contrast, achieves the highest
marginal binding-active rate (27.16\%), but only 34.09\% of its submitted
candidates satisfy the binding-free Safe-only criterion.

Pepti-drift remains competitive on both axes. Its 61.17\%
Safe-only rate is second only to PepTune, and it ranks second in toxicity,
hemolysis, and neurotoxicity pass rates. On the binding axis, Pepti-drift
achieves the highest confidence-gated binding coverage (35.45\%) and target
pass rate (26.14\%), while its marginal binding-active rate (25.80\%) is
second only to PepMLM.

Taken together, the marginal results address \textbf{(Q2)}: Pepti-drift's joint performance is not driven by strength on only one marginal axis.
However, marginal performance alone does not establish that the two properties are satisfied by the \emph{same peptide}.
We therefore next return to the primary fixed-budget \safeactive{} endpoint to evaluate joint multi-property success.





\subsection{Pepti-drift shifts the joint quality--cost frontier}
\label{sec:jointresults}


To answer \textbf{(Q3)}, we evaluate whether Pepti-drift achieves competitive fixed-budget \safeactive{} performance without the computational overhead of inference-time guidance.

\begin{figure*}[t]
\centering
\includegraphics[width=0.98\linewidth]
{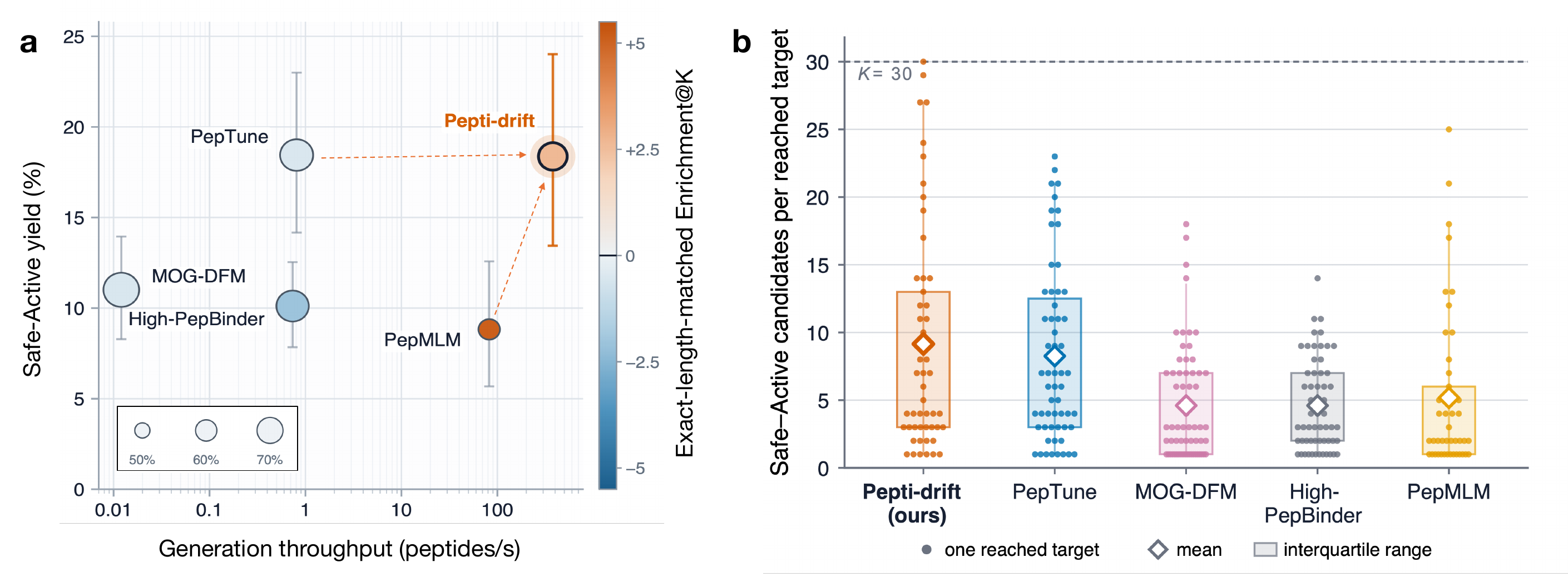}
\vspace{-1em}
\caption{
\textbf{Pepti-drift combines efficient joint generation with deep target-level discovery.}
\textbf{(a)} \safeactive{} yield versus generation throughput at $K=30$.
Marker area denotes target coverage, color denotes exact-length-matched
Enrichment@30, and error bars show target-bootstrap 95\% confidence intervals.
\textbf{(b)} Discovery depth among reached targets.
Points denote individual targets, diamonds the mean, and boxes the
interquartile range; the dashed line marks the $K=30$ budget.
}
\label{fig:quality_cost_depth}
\end{figure*}

Pepti-drift achieves an 18.37\% \safeactive{} yield, only 0.08 percentage points below PepTune's highest raw yield of 18.45\%, while requiring only 2.643 ms per peptide (Appendix~\ref{sec:full_joint_results}). This corresponds to a $468\times$ lower measured generation cost than PepTune. 
This result is best understood through the joint quality--cost frontier (Figure~\ref{fig:quality_cost_depth}a).

Existing methods generally trade throughput for joint candidate quality:
the AR Transformer is fast but achieves lower yield, whereas PepTune achieves the highest raw yield at substantial cost. Pepti-drift shifts this frontier by combining near-best absolute joint yield, $+2.65$ positive exact-length-matched enrichment, and fast generation. Although no method dominates every metric, Pepti-drift is the only one to combine these three properties.
Compared to a speed-matched autoregressive Transformer control (2.940\,ms per peptide), Pepti-drift generates candidates slightly faster (2.643\,ms) while improving raw \safeactive{} yield by 3.82 percentage points, suggesting that the full Pepti-drift stack improves the performance gain. Furthermore, these efficiently generated outputs strictly satisfy the benchmark's sequence-integrity requirements: they are 100\% valid and unique within targets, 99.51\% exact-novel relative to training peptides, and exhibit no exact cross-target reuse (Appendix~\ref{app:fixedbudget}).
These findings answer \textbf{(Q3)}: Pepti-drift achieves near-best joint \safeactive{} yield at a fraction of the generation cost of inference-guided methods, without sacrificing sequence validity or novelty.


\subsection{Pepti-drift yields deeper candidate pools on reached targets}
\label{sec:jointdiscovery}


Finding one \safeactive{} candidate for a target is only a starting point:
multiple distinct candidates provide more options for downstream testing and optimization.
To answer \textbf{(Q4)}, we distinguish \emph{target coverage},
the fraction of targets with at least one \safeactive{} candidate, from \emph{candidate count per reached target}, the number of such candidates among the $K=30$ submissions for each target.


Pepti-drift averages 9.15 \safeactive{} candidates per reached target,
compared with 8.25 for PepTune and approximately 4.6--5.2 for the other
published methods (Figure~\ref{fig:quality_cost_depth}b). Thus, it provides
about 11\% more candidates per reached target than PepTune under the same
sampling budget, at $468\times$ lower generation cost. This advantage extends
to targets with many successful candidates: Pepti-drift produces at least 21
\safeactive{} candidates on seven targets and is the only compared method
to produce at least 26 on any target, including all $30/30$ on one target
(Appendix~\ref{app:discovery_depth}).


These averages are conditional on each method's own reached targets and
complement target coverage (Figure~\ref{fig:quality_cost_depth}a). Together, these results answer \textbf{(Q4)}: conditional on reaching a target, Pepti-drift combines low generation cost with more candidates for
follow-up on its reached targets. We next examine the predicted structures on targets where Pepti-drift provides the largest gains in candidate count.

\subsection{Pepti-drift exhibits physically grounded interface formation}
\label{sec:structural_analysis}

Optimizing proxy scores raises a familiar ML concern:
\emph{reward hacking}, where models exploit evaluator imperfections to improve scores without improving the intended properties.
To address \textbf{(Q5)}, we look beyond scalar scores and examine whether Pepti-drift's successful candidates exhibit plausible target interactions in their predicted structures.
We quantify contact density, buried solvent-accessible surface area, and hydrophobic side-chain burial, normalizing contact counts and surface areas by peptide length \citep{chen2013interface,munteanu2015solvent}.

\begin{figure*}[t]
\centering

\includegraphics[width=0.497\textwidth]
{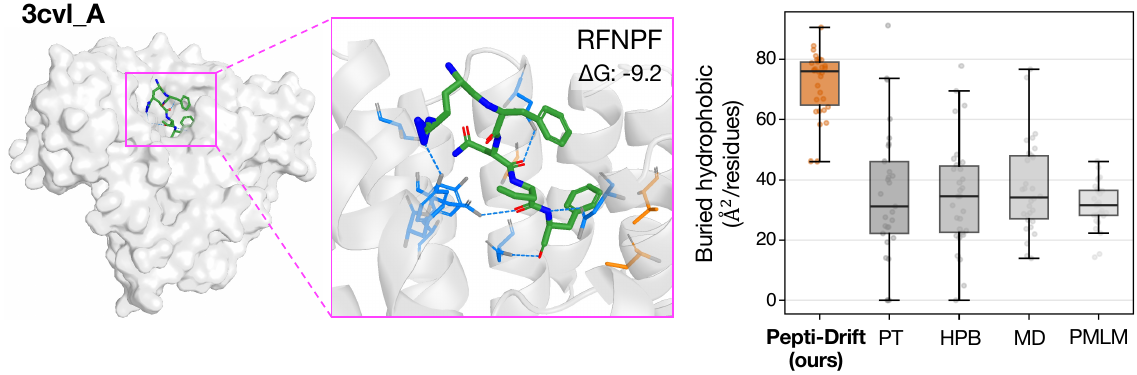}
\hfill
\includegraphics[width=0.497\textwidth]
{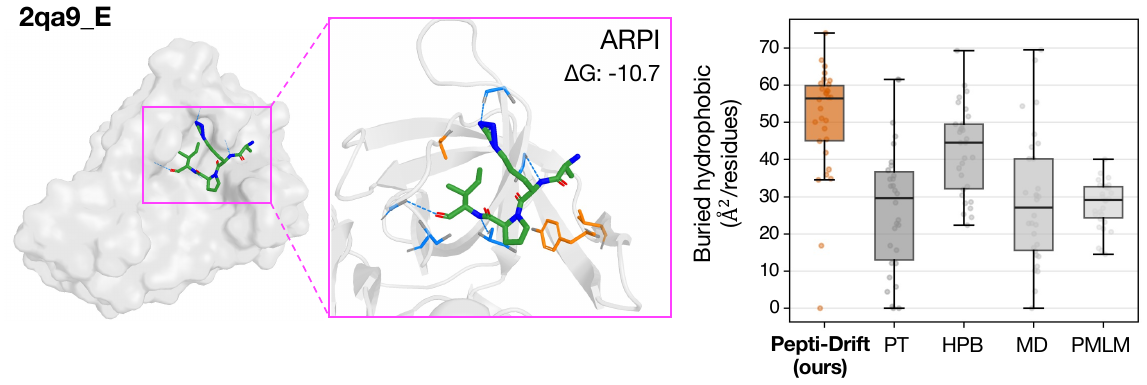}

\caption{
\textbf{Predicted interfaces and hydrophobic burial for two high-depth \safeactive{} targets.}
Representative Pepti-drift complexes and enlarged target--peptide interfaces are shown for 3cvl\_A (left; RFNPF) and 2qa9\_E (right; ARPI), together with buried hydrophobic surface area per peptide residue across methods.
Predicted $\Delta G$ is reported in kcal/mol.
Peptides are shown in green; polar- and hydrophobic-contacting target residues in light blue and orange, respectively; blue dashed lines indicate putative polar contacts.
Boxes show the interquartile range and median; points show individual candidates.
PT, PepTune; HPB, High-PepBinder; MD, MOG-DFM; PMLM, PepMLM.
}

\label{fig:structural_interpretation}

\end{figure*}

We analyze 3cvl\_A and 2qa9\_E, selected by the prespecified rule for the
largest candidate-count gains among targets reached by all five methods
(Appendix~\ref{app:target_examples}). Each method's representative is chosen
by maximum predicted interface confidence (ipTM). On 3cvl\_A, Pepti-drift's
representative has the highest contact density and largest buried area per
residue among the five representatives, with 79\% of its hydrophobic
side-chain surface buried against the target. 
Moreover,
the candidate distributions show greater median buried hydrophobic surface
per residue for Pepti-drift on both targets
(Figure~\ref{fig:structural_interpretation}).
On 2qa9\_E, Pepti-drift's representative is at least as hydrophobic at the
sequence level as MOG-DFM's, yet has a lower predicted hemolysis score (0.119 versus 0.902). Its predicted structure also buries a larger fraction of hydrophobic surface and leaves less exposed per residue. Thus, it 
combines favorable predicted binding and safety with substantial
interface burial, without simply reducing sequence hydrophobicity.
Full cross-method descriptors are reported in Appendix
Table~\ref{tab:structural_high_depth}.
Similar interface-utilization patterns are also observed on two high-depth PepTune targets, indicating that the trend extends beyond the Pepti-drift-selected targets (Appendix Figures~\ref{fig:contact_anal_pepti_drift} and~\ref{fig:contact_anal_peptune}).

Overall, these structural analyses answer \textbf{(Q5)}: Pepti-drift's joint success is more consistent with physically grounded 3D interface formation than with simple 1D compositional biases.
The analyses are interpretive rather than independent validation and do not rule out all forms of reward hacking.


\section{Conclusion}
Before concluding, we acknowledge several limitations.
This study remains entirely \textit{in silico}: predicted binding and safety require prospective \emph{in vitro} and \emph{in vivo} validation to establish biological viability. Moreover, \benchmark{} captures only part of peptide developability, and future work should incorporate stability, pharmacokinetics, bioavailability, and immunogenicity.

Our component analysis also indicates that removing the drifting block produces only a modest reduction in joint performance with substantially overlapping confidence intervals in Appendix \ref{app:component_analysis}. 
Thus, the present results support the effectiveness of Pepti-drift's full one-step generation framework, indicating that the performance gains mainly stem from the architecture as a whole rather than being dominated by the drifting block.
Developing training objectives and latent-space control mechanisms that make attraction--repulsion refinement more influential is an important direction for future work.


In conclusion, we introduced \benchmark{}, revealing that the apparent negative association between predicted binding and safety is largely explained by peptide length. This finding motivated our use of exact-length-matched controls for fair joint evaluation. To this end, we proposed Pepti-drift, a one-step peptide generation framework that achieves competitive joint binding--safety control by amortizing it into the training phase, thereby avoiding costly repetitive guidance at inference time.

In our experiments, we demonstrated that shifting multi-property control from iterative inference to learned one-step refinement fundamentally improves the quality--cost trade-off of peptide generation. Specifically, Pepti-drift achieves an 18.37\% \safeactive{} yield with a +2.65 percentage-point exact-length-matched enrichment. At just 2.643 ms per peptide, it incurs a 468$\times$ lower generation cost than the strongest inference-guided baseline. Overall, these results establish a more rigorous paradigm for joint evaluation and prove that competitive multi-property generation is highly scalable without costly inference-time guidance.

\section{AI Use Statement}
\label{app:ai_use_statement}
Generative AI tools were used to assist with manuscript organization,
language editing, and presentation of author-provided experimental results.
They were not used to generate the benchmark measurements reported in this
work. All AI-assisted text and interpretations were reviewed by the authors,
numerical claims were checked against saved machine-readable outputs, and the
cited literature was manually verified. The authors take responsibility for
the final text, claims, analyses, and released artifacts.

\section{Ethics Statement}
\label{app:ethics}

This work uses public protein, peptide, and assay-derived datasets and does not
involve human participants or new animal experiments. Generative peptide models
are dual-use: improved design may accelerate therapeutic discovery, but could
also be used to seek toxic or otherwise harmful sequences. The benchmark
reports computational predictions only and must not be used to claim biological
safety or to bypass experimental biosafety review. Synthesis and testing should
follow institutional, legal, and assay-specific procedures. Released sequences,
models, and evaluator outputs should carry provenance, licenses, intended-use
constraints, and explicit warnings about predictor uncertainty.



\section{Reproducibility Statement}
\label{app:reproducibility_statement}
The source code for Pepti-drift and BindSafe-PepBench will be made available upon acceptance of the paper, and data subject to non-shareable licenses are fully described in the Appendix.
We provide the information required to reproduce the model, benchmark, and
evaluation pipeline throughout the Appendix. Pepti-drift model definitions and
training objectives are given in Appendix~\ref{app:model}, and the benchmark
protocol, datasets, leakage safeguards, and evaluators are described in
Appendix~\ref{app:metrics}. Robustness analyses and additional fixed-budget
diagnostics are reported in Appendix~\ref{app:tradeoff}
and~\ref{app:fixedbudget}, respectively. Architecture and training
hyperparameters are summarized in Appendix~\ref{app:model_hyperparameters},
while comparator configurations, pretrained checkpoints, and generation-speed
protocols are provided in Appendix~\ref{app:reproducibility}.

\section{Acknowledgments}
Figures 1, 2, and 6 were created with BioRender (https://BioRender.com). We also thank the authors of PepBench, PepBenchmark, PepMLM, PepTune, MOG-DFM, and High-PepBinder for publicly releasing their datasets and model checkpoints, which made the comparisons in BindSafe-PepBench possible.

\bibliographystyle{iclr2027_conference}
\bibliography{reference_v2}


\appendix

\section{Pepti-drift Implementation Details}
\label{app:model}
\subsection{Training data}

Binding-positive examples were obtained from the official PepBench splits
\citep{kong2024pepbenchdata}.  After removing 10 training rows whose peptide
matched one of five held-out test peptides, the exported train/validation/test
sets contained 4,060/105/88 protein--peptide pairs.  Deduplication of identical
protein-sequence--peptide-sequence tuples by the training loader resulted in
3,953/104/88 effective pairs, respectively.  For negative repulsion, we used
canonical 1--50-residue peptides labeled toxic and/or hemolytic in the
PepBenchmark training split (hybrid split, seed 0)
\citep{zhang2026pepbenchmark}.  Deduplication produced 3,761 sequences; removing
184 sequences overlapping the property test set and excluding all exact
binding-positive overlaps left 3,577 negatives.  A deterministic 90/10 split
(seed 42) assigned 3,219 negatives to training and 358 to validation.  The
held-out test set was used neither for optimization nor checkpoint selection.

\subsection{Trainable model and fixed peptide reference map}
\label{app:model_representation}

We distinguish the parameters optimized by Pepti-drift from the fixed
components used to construct its training targets.

Let
\begin{equation}
\Theta
=
\{
\theta_A,
\theta_G,
\theta_D,
\theta_{\mathrm{dec}}
\}
\end{equation}
denote the trainable parameters of the target projection, conditional latent
generator, one-step drift module, and non-autoregressive decoder, respectively.

Target protein and peptide sequences are first represented by frozen, mean-pooled
ESM-2 3B embeddings,
\begin{align}
e_A(a)
&\in\mathbb{R}^{2560},\\
e_P(y)
&\in\mathbb{R}^{2560}.
\end{align}

The target embedding is mapped through a trainable two-layer projection,
\begin{equation}
h_{\theta_A}(a)
=
\phi_{A,\theta_A}(e_A(a))
\in\mathbb{R}^{2560}.
\end{equation}
This projection lies on the differentiable path to the generator, drift
module, and decoder and is optimized jointly with the remainder of the model.

Peptide references are instead mapped by a separate fixed two-layer MLP,
\begin{align}
r_0(y)
&=
\phi_{P,0}(e_P(y))
\in\mathbb{R}^{d_z},\\
u_0(y)
&=
\operatorname{norm}(r_0(y))
=
\frac{r_0(y)}
     {\lVert r_0(y)\rVert_2}
\in\mathbb{S}^{d_z-1},
\qquad d_z=1024.
\end{align}
The map $\phi_{P,0}$ has architecture
\begin{equation}
2560
\rightarrow
1024
\rightarrow
\operatorname{ReLU}
\rightarrow
1024.
\end{equation}
Its parameters are randomly initialized once under the model initialization
seed and are held fixed at those initial values throughout Pepti-drift
training. In particular, $\phi_{P,0}$ is not pretrained and is not optimized
by any loss in this work.

The purpose of $\phi_{P,0}$ is therefore only to provide a stationary
$d_z$-dimensional reference coordinate system in which positive and
safety-associated peptide examples can be compared with generated latents.
We do not interpret this projection itself as a learned binding- or
safety-specific representation.

For each positive training pair $(a_i,y_i^+)$, the corresponding attractive
reference is
\begin{equation}
u_i^+
=
u_0(y_i^+).
\end{equation}
The same map is applied to all safety-associated peptide references,
\begin{equation}
u_j^-
=
u_0(y_j^-).
\end{equation}

Because all reference vectors are $\ell_2$-normalized,
\begin{equation}
\lVert u-v\rVert_2^2
=
2-2u^\top v,
\qquad
u,v\in\mathbb{S}^{d_z-1},
\end{equation}
so ranking normalized references by squared Euclidean distance is equivalent
to ranking them by cosine similarity.

\subsection{One-step pushforward and detached target construction}

For each training example $i$, we sample
\begin{equation}
\xi_i
\sim
\mathcal{N}(0,I_{d_z})
\end{equation}
and compute
\begin{align}
z_{i,0}
&=
G_{\theta_G}
\left(
h_{\theta_A}(a_i),\xi_i
\right),\\
\Delta_{i,0}
&=
D_{\theta_D}
\left(
z_{i,0},h_{\theta_A}(a_i)
\right),\\
x_i=z_{i,1}
&=
z_{i,0}
+
\eta\Delta_{i,0},
\qquad \eta=1.
\end{align}

Equivalently,
\begin{equation}
x_i
=
g_\Theta(a_i,\xi_i),
\end{equation}
with conditional pushforward distribution
\begin{equation}
q_{\Theta,a_i}
=
\left(
g_\Theta(a_i,\cdot)
\right)_\#
\mathcal{N}(0,I_{d_z}).
\end{equation}

We normalize the refined latent,
\begin{equation}
\bar{x}_i
=
\operatorname{norm}(x_i).
\end{equation}

Hard-negative selection is performed using a detached copy of the current
latent. For each available negative reference,
\begin{equation}
d_{ij}^-
=
\frac{1}{d_z}
\left\|
\operatorname{sg}(\bar{x}_i)
-
u_{ij}^-
\right\|_2^2 .
\end{equation}
Let $\mathcal{H}_i$ contain the indices of the
$K_{\mathrm{neg}}$ smallest distances. For $j\in\mathcal{H}_i$,
\begin{equation}
w_{ij}^-
=
\frac{
\exp(-d_{ij}^-/\tau)
}{
\sum_{\ell\in\mathcal{H}_i}
\exp(-d_{i\ell}^-/\tau)
}.
\end{equation}

We expand the attraction--repulsion field for training example $i$ at epoch $e$ as
\begin{equation}
V_i(\bar{x}_i;e)
=
\alpha_+ V_{\text{attract}}(\bar{x}_i)
+
\alpha_-(e) V_{\text{repel}}(\bar{x}_i),
\end{equation}
where $\alpha_+=1$. Specifically,
\begin{equation}
V_i(\bar{x}_i;e)
=
\left(
u_i^+-\bar{x}_i
\right)
+
\alpha_-(e)
\sum_{j\in\mathcal{H}_i}
w_{ij}^-
\left(
\bar{x}_i-u_{ij}^-
\right),
\end{equation}
where the repulsion coefficient is linearly warmed up as
\begin{equation}
\alpha_-(e)
=
\alpha_{-}^{\mathrm{max}}
\min\left(
1,\,
\frac{\max(0,e-e_{\mathrm{start}})}
{e_{\mathrm{warmup}}}
\right).
\end{equation}
Here, $\alpha_{-}^{\mathrm{max}}$ denotes the final repulsion strength,
$e_{\mathrm{start}}$ the epoch at which repulsion is activated, and
$e_{\mathrm{warmup}}$ the number of epochs over which $\alpha_-(e)$
increases linearly from zero to $\alpha_{-}^{\mathrm{max}}$.

The detached alignment target is
\begin{equation}
\widetilde{x}_i
=
\operatorname{sg}
\left[
\operatorname{norm}
\left(
\bar{x}_i+V_i(\bar{x}_i;e)
\right)
\right].
\end{equation}

The stop-gradient makes target construction asymmetric: $\widetilde{x}_i$
is recomputed from the current forward-pass latent but is treated as fixed
during the corresponding backward pass. Consequently, the drift loss updates
the generator and drift module through $\bar{x}_i$, but does not update the
positive reference map, negative reference map, hard-negative selection, or
kernel weights.

\subsection{Non-causal decoder and optimization objective}

Let $L=50$ be the maximum decoder length. For position
$t\in\{1,\ldots,L\}$, the initial decoder representation is
\begin{equation}
H_{i,t}^{(0)}
=
q_t
+
W_zx_i
+
W_ah_{\theta_A}(a_i),
\end{equation}
where $q_t$ is a learned positional query. A Transformer encoder without a
causal mask processes all positions jointly,
\begin{equation}
H_i
=
\mathcal{T}_{\theta_{\mathrm{dec}}}
\left(
H_i^{(0)}
\right),
\end{equation}
followed by a token projection and softmax,
\begin{align}
\ell_{i,t}
&=
W_oH_{i,t}+b_o,\\
p_{i,t}
&=
\operatorname{softmax}(\ell_{i,t}).
\end{align}

The resulting conditional factorization is
\begin{equation}
P_\Theta(s_{i,1:L}\mid a_i,\xi_i)
=
\prod_{t=1}^{L}
P_\Theta
\left(
s_{i,t}\mid
x_i,h_{\theta_A}(a_i)
\right).
\end{equation}
Although the distribution factorizes across positions, their representations interact through non-causal self-attention before token prediction. All positions are therefore decoded in one parallel pass. The generated sequence is truncated at the first \texttt{<eos>} token output; thus, the length is not supplied explicitly and $L=50$ specifies only the maximum decoded length.
Let
\begin{equation}
\Omega
=
\{
(i,t):
s_{i,t}^+\neq\texttt{<pad>}
\}
\end{equation}
denote the non-padding target positions. The sequence loss is
\begin{equation}
\mathcal{L}_{\mathrm{seq}}
=
-\frac{1}{|\Omega|}
\sum_{(i,t)\in\Omega}
\log
P_\Theta
\left(
s_{i,t}=s_{i,t}^+
\mid
a_i,\xi_i
\right).
\end{equation}

The drift-alignment loss is
\begin{equation}
\mathcal{L}_{\mathrm{drift}} = \frac{1}{Bd_z} \sum_{i=1}^{B} \left\| \operatorname{norm}(x_i)-\widetilde{x}_i \right\|_2^2.
\end{equation}
We have
$
\left\| \operatorname{norm}(x_i)-\widetilde{x}_i \right\|_2^2 = 2-2\operatorname{norm}(x_i)^\top\widetilde{x}_i,
$
and thus, minimizing $\mathcal{L}_{\mathrm{drift}}$ is equivalent to maximizing the cosine similarity between the refined latent and the attraction--repulsion target.
The displacement regularizer is
\begin{equation}
\mathcal{L}_{\mathrm{reg}}
=
\frac{1}{Bd_z}
\sum_{i=1}^{B}
\left\|
\Delta_{i,0}
\right\|_2^2.
\end{equation}

The complete objective is
\begin{equation}
\mathcal{L}_{\mathrm{total}}
=
\lambda_{\mathrm{seq}}
\mathcal{L}_{\mathrm{seq}}
+
\lambda_{\mathrm{drift}}
\mathcal{L}_{\mathrm{drift}}
+
\lambda_{\mathrm{reg}}
\mathcal{L}_{\mathrm{reg}}
.
\end{equation}

Optimization is performed over the trainable model parameters $\Theta$,
\begin{equation}
\Theta^\star
=
\operatorname*{arg\,min}_{\Theta}
\;
\mathbb{E}
\left[
\mathcal{L}_{\mathrm{total}}
\right].
\end{equation}
The ESM-2 encoders and peptide reference map are fixed:
\begin{equation}
\nabla_{\omega_{\mathrm{ESM}}}
\mathcal{L}_{\mathrm{total}}
=
0,
\qquad
\nabla_{\phi_{P,0}}
\mathcal{L}_{\mathrm{total}}
=
0.
\end{equation}
In contrast,
\begin{equation}
\nabla_{\theta_A}
\mathcal{L}_{\mathrm{total}},
\quad
\nabla_{\theta_G}
\mathcal{L}_{\mathrm{total}},
\quad
\nabla_{\theta_D}
\mathcal{L}_{\mathrm{total}},
\quad
\nabla_{\theta_{\mathrm{dec}}}
\mathcal{L}_{\mathrm{total}}
\end{equation}
are generally nonzero.

At inference, the target peptide, safety-derived negative references,
$\phi_{P,0}$, and the attraction--repulsion target are not evaluated.
Generation requires only the target protein representation, sampled Gaussian
noise, one generator evaluation, one learned drift update, and one parallel
decoder evaluation.

\subsection{Component analysis}
\label{app:component_analysis}

We additionally evaluate several components of Pepti-drift using an
independently frozen seed-45 run (Table~\ref{tab:ablation_seed45}).
The full model achieves the highest \safeactive{} rate (17.80\%) and
exact-length-matched enrichment ($+7.54$ percentage points).
Removing the drifting block results in only a modest reduction, whereas
removing the one-step Transformer or replacing ESM-2 representations with
one-hot features progressively reduces joint performance. Because this
analysis uses a single frozen training seed and the confidence intervals
overlap substantially, we interpret these results as descriptive
component-level evidence rather than statistically separable causal effects
of individual components.

\begin{table}[htb]
\setlength{\belowcaptionskip}{4pt}
\caption{
\textbf{Component analysis of Pepti-drift at $K=30$.}
Results are from an independently frozen seed-45 run.
Enrichment is the percentage-point improvement over the target- and
exact-length-matched random control; brackets show target-bootstrap 95\%
confidence intervals.
}
\label{tab:ablation_seed45}
\centering
\scriptsize
\setlength{\tabcolsep}{4.0pt}
\begin{tabular}{lcc}
\toprule
Condition &
SA Enrichment $\uparrow$ &
SA rate $\uparrow$ \\
\midrule

Full
& $\mathbf{+7.54}$ [3.33, 12.27]
& \textbf{17.80\% [12.54, 23.33]} \\

No drifting block
& $+7.05$ [2.50, 11.63]
& 17.31\% [12.12, 22.84] \\

No one-step Transformer
& $+3.90$ [0.68, 7.54]
& 14.17\% [10.00, 18.56] \\

No ESM-2 (one-hot)
& $+2.35$ [$-1.02$, 5.91]
& 12.61\% [8.71, 17.27] \\

\bottomrule
\end{tabular}
\end{table}

\section{BindSafe-PepBench Details}
\label{app:metrics}

\subsection{Task, endpoint, and matched control}

\begin{figure}[t]
    \centering
    \includegraphics[width=0.90\textwidth]
    {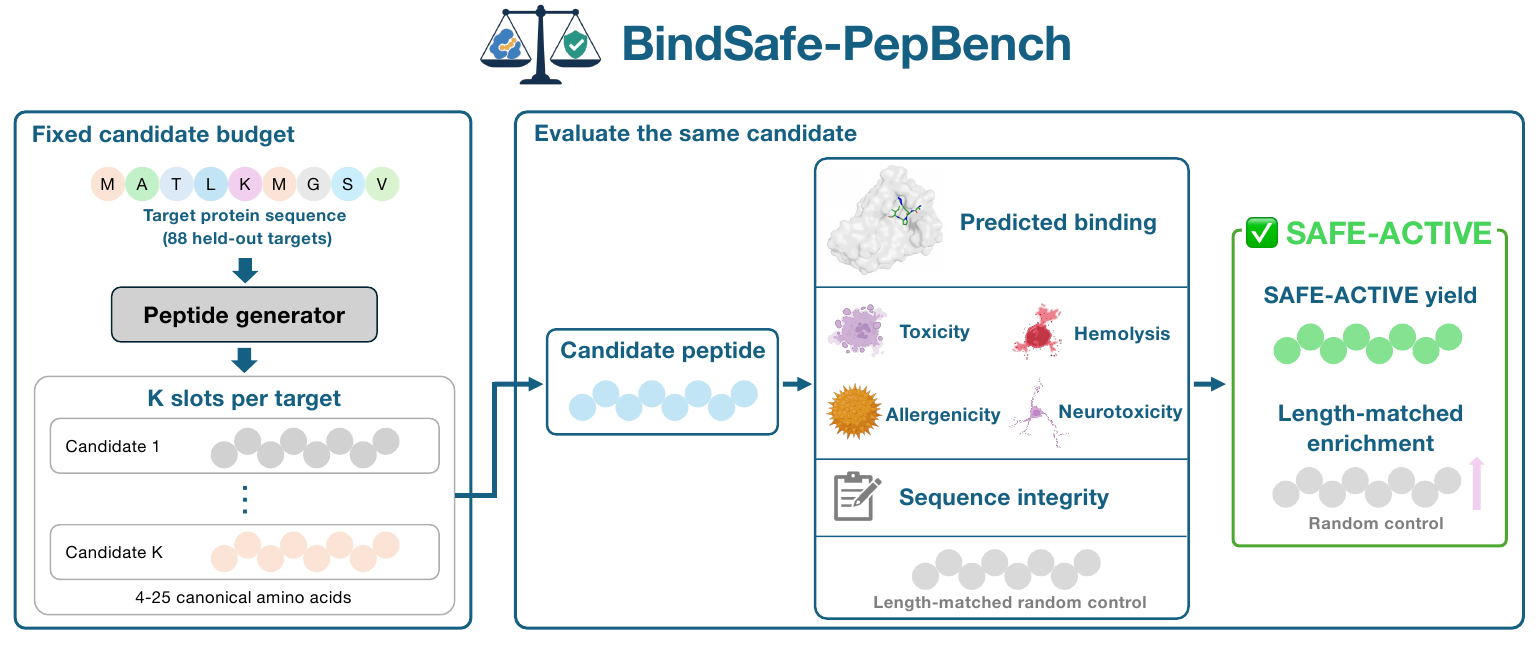}
    \caption{
    \textbf{Overview of BindSafe-PepBench.}
    Given only a target protein sequence, each method generates a fixed budget of $K$ candidate peptides for each of 88 held-out targets.
    The same candidate is jointly evaluated for predicted target binding, four safety liabilities (toxicity, hemolysis, allergenicity, and neurotoxicity), and sequence integrity (validity, uniqueness, novelty, reuse).
    Candidates satisfying all criteria are counted as \safeactive{}.
    The benchmark reports raw \safeactive{} yield together with exact-length-matched enrichment against random peptides matched to the same target and peptide length.
    }
    \label{fig:bindsafe_overview}
\end{figure}

For each target protein sequence $a_t$, a submission provides an ordered list of
up to $K=30$ peptide candidates. The primary track restricts peptides to
4--25 canonical amino acids. Every submitted slot remains part of the fixed
budget, so invalid, duplicated, out-of-range, underfilled, or unevaluable slots
cannot be removed after generation (Figure \ref{fig:bindsafe_overview}).

A candidate is \safeactive{} only if it satisfies the common structure and
binding criteria
\[
\mathrm{ipTM}\geq0.5,\qquad
\mathrm{interface\ pLDDT}\geq70,\qquad
\bindscore\geq7,
\]
and all four safety scores---toxicity, hemolysis, allergenicity, and
neurotoxicity---are at most 0.5. It must additionally be canonical, have length
4--25, be an exact non-match to generator training peptides, be the first
occurrence within its target, and be reused exactly across at most two targets.

Raw \safeactive{} yield is the primary metric and uses the full
allocated candidate budget as its denominator.
Exact-length-matched Enrichment@$K$ is reported alongside it as a key
complementary metric.
The two measures are interpreted jointly to distinguish absolute
joint success from improvement over a matched random baseline.

For Enrichment@$K$, every observed candidate is paired with a uniformly sampled
random peptide that preserves its target, submitted rank, and exact peptide
length. Because peptide length can affect both binding and safety
evaluator outputs, this matching prevents Enrichment@$K$ from rewarding methods
merely for producing a favorable length distribution. Enrichment is computed as
the target-macro difference in \safeactive{} rate between the method and its
paired control. Percentile intervals use 2,000 target-bootstrap replicates with
seed 0.

The matched control therefore controls for target and exact peptide length, but does not explicitly match amino-acid composition, net charge, hydrophobicity, or
aromatic content. Enrichment@$K$ should consequently be interpreted as
improvement over random sequences of the same length, rather than over a fully
composition-matched generator.

\subsection{Dataset construction and leakage safeguards}
\label{app:data}

The binding dataset is constructed from the official PepBench split and contains
4,263 target--peptide complexes: 4,070 training, 105 validation, and 88 held-out
test examples~\citep{kong2024pepbenchdata}. The identifiers correspond to 2,886
unique target sequences, and exact target sequences do not cross from training
to validation or test. Held-out native peptides have lengths 4--21 with median
10.

For trainable generators, 10 training rows whose peptide sequence exactly
matches one of five held-out test peptides are removed, leaving 4,060 binding
pairs. The safety-derived repulsive pool starts from 3,761 unique positive
training-split property peptides. Removing 184 sequences that occur in the
property test set leaves 3,577 sequences, with zero exact overlap with the
property test set or with any binding-positive peptide. A total of 162 sequences
also occur in the property validation split; these validation labels are not
used to evaluate generated candidates.

At the target level, the held-out split is also dissimilar from training under
a simple sequence-overlap audit: the median held-out target shares 5.0\% of its
4-mers with its most similar training target, while 4.5\% and 2.3\% of targets
exceed 30\% and 50\% containment, respectively. For short 4--25-residue
peptides, nearest-neighbor sequence identity has a high chance floor and is
therefore not used as a novelty threshold. The benchmark instead applies exact
sequence novelty and exact reuse constraints.

\subsection{Property evaluator performance}
\label{app:evaluators}

Table~\ref{tab:evaluators} reports held-out performance of the four frozen
property models that define the \safeactive{} endpoint. The positive class is
the unsafe class for every task. Benchmark gating uses a common score threshold
of 0.5; this threshold is intended as a consistent benchmark decision rule
rather than as a calibrated biological risk boundary.

\begin{table}[h]
\setlength{\belowcaptionskip}{4pt}
\caption{Held-out performance of the four property evaluators used in the
\safeactive{} endpoint. BA is balanced accuracy and MCC is Matthews
correlation coefficient.}
\label{tab:evaluators}
\centering
\small
\begin{tabular}{lrrrrr}
\toprule
Property & Test $n$ & ROC-AUC & PR-AUC & BA & MCC \\
\midrule
Toxicity       & 407 & 0.756 & 0.628 & 0.663 & 0.344 \\
Hemolysis      & 432 & 0.825 & 0.674 & 0.733 & 0.462 \\
Allergenicity  & 255 & 0.874 & 0.758 & 0.759 & 0.533 \\
Neurotoxicity  & 317 & 0.700 & 0.582 & 0.668 & 0.329 \\
\bottomrule
\end{tabular}
\end{table}

\section{Robustness of Binding--Safety Evaluation}
\label{app:tradeoff}

Because the central binding--safety finding depends on computational models
on both axes, it is important to distinguish conclusions that persist across
evaluator choices from those specific to a particular oracle. We therefore
audit the observed association in three steps. We first characterize the binding--safety association under the primary benchmark evaluators and examine whether it is driven by particular methods or targets. We then vary the binding oracle while holding safety fixed, and finally vary the safety oracle while keeping the generated candidate sets fixed. Together, these analyses provide a two-sided stress test of the
computational evaluation pipeline. They are intended to assess robustness of
the benchmark conclusions, not to replace experimental validation.

\subsection{Primary association and method-level diagnostics}

We first establish the binding--safety relationship under the primary benchmark
evaluators. The descriptive analysis retains valid, in-range, nonduplicate
candidates with complete binding and safety scores. Binding and safety are
ranked separately within each method--target stratum before pooling, preventing
between-method or between-target mean shifts from inducing the reported
association. Confidence intervals use 20,000 target-cluster bootstrap replicates.
Target-specific permutation tests use 20,000 permutations followed by
Benjamini--Hochberg correction~\citep{benjamini1995controlling}.

The primary safety utility uses only the four liabilities defining the
\safeactive{} endpoint. Across 6,568 complete candidates from 12 submissions,
the stratified binding--safety association is negative
($\rho=-0.224$, 95\% CI $-0.253$ to $-0.195$), with a negative within-target
association for 73 of 82 estimable targets. Toxicity shows the strongest
component-level association with binding ($\rho=-0.277$).

An auxiliary utility that additionally includes nonfouling yields a weaker but
still negative aggregate association
($\rho=-0.131$, 95\% CI $-0.160$ to $-0.098$). We therefore focus the main
analysis on the four liabilities that directly define the benchmark endpoint.

\begin{figure}[h]
\centering
\includegraphics[width=\linewidth]
{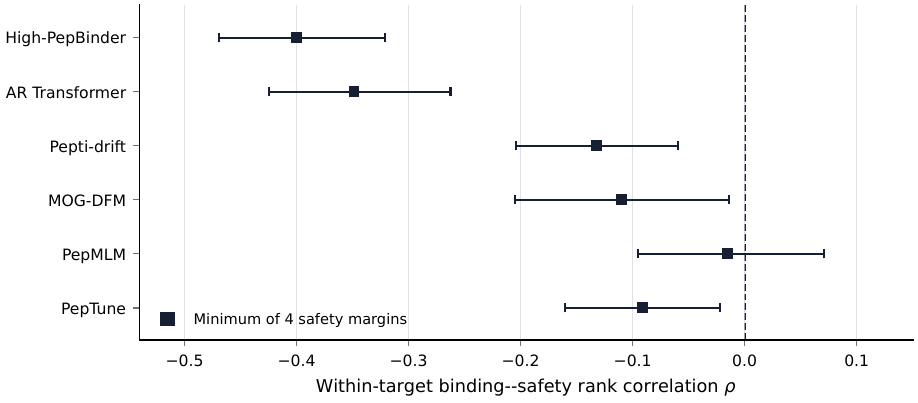}
\caption{
Method-specific within-target predicted binding--safety associations for the
principal learned or guided methods. Points are Spearman estimates and bars are
target-bootstrap 95\% confidence intervals. The safety score is the minimum of
the four liability margins defining the \safeactive{} endpoint, excluding the
auxiliary nonfouling margin. The dashed line denotes zero association.
}
\label{fig:tradeoff_by_method}
\end{figure}

Figure~\ref{fig:tradeoff_by_method} shows that the pooled result is not driven
by a single generator. Under the four-liability utility, the association is
negative for every principal learned or guided method, although its magnitude
varies across methods and individual method-level estimates need not all differ
from zero. Estimates range from $\rho=-0.400$ for High-PepBinder
(95\% CI $-0.469$ to $-0.321$) to $\rho=-0.015$ for PepMLM
($-0.095$ to $0.071$), the only method whose interval spans zero.

\subsection{Robustness to independent binding oracles}
\label{app:binding_oracles}

Having established the association under the primary benchmark pipeline, we
first test whether its direction depends on the binding oracle. The primary
benchmark uses ESMFold2 together with PRODIGY. We therefore add two
independent scoring tracks: OpenDDE+EvoEF2 and OpenDDE+HADDOCK3
\citep{aureka2026opendde,huang2020evoef2,giulini2025haddock3}. The latter two
share an OpenDDE pose but use different scoring functions, whereas the primary
pipeline differs in both structure predictor and scorer.

All scores are oriented so that larger values indicate more favorable predicted
binding. Because the three scoring functions have different units and
calibration, we compare continuous within-method--target ranks and do not
transfer the PRODIGY activity threshold across oracles. The OpenDDE panel
contains five learned or guided methods and the uniform-random control over all
88 targets. OpenDDE produced 15,834/15,834 requested poses, and both EvoEF2 and
HADDOCK3 returned scores for every pose; 15,596 candidates remain after the
common safety and integrity join.

Table~\ref{tab:oracle_robustness} reports the four-liability analysis used in
the main benchmark. These values are not expected to exactly match the primary
complete-case statistic above: this analysis uses continuous binding scores
without the benchmark's binary structure-confidence and binding gates, and the
candidate set differs across oracle panels. Despite these differences, all three
target-macro confidence intervals exclude zero, preserving the negative
aggregate direction under substantially different binding pipelines.

\begin{table}[htb]
\centering
\setlength{\belowcaptionskip}{4pt}
\caption{
Binding--safety association under three binding oracles. The safety utility is
the minimum signed margin over the four liabilities defining the \safeactive{}
endpoint. Intervals are target-bootstrap 95\% confidence intervals.
Cross-oracle differences in correlation magnitude are not interpreted as
differences in oracle quality.
}
\label{tab:oracle_robustness}
\small
\setlength{\tabcolsep}{4pt}
\begin{tabular}{@{}lrrrr@{}}
\toprule
Oracle &
$n$ &
Strat. $\rho$ &
\shortstack{Target-macro $\rho$\\{[95\% CI]}} &
\shortstack{Negative\\targets} \\
\midrule
ESMFold2+PRODIGY
& 23,993
& $-0.235$
& $-0.235$ [$-0.250$, $-0.219$]
& 100.0\% \\
OpenDDE+EvoEF2
& 15,596
& $-0.098$
& $-0.098$ [$-0.117$, $-0.078$]
& 85.2\% \\
OpenDDE+HADDOCK3
& 15,596
& $-0.137$
& $-0.137$ [$-0.154$, $-0.120$]
& 94.3\% \\
\bottomrule
\end{tabular}
\end{table}

Table~\ref{tab:oracle_method_rho} resolves the same quantity by method. The
association is negative for every method under all three oracles with one
exception: PepTune, whose estimate is close to zero under
ESMFold2+PRODIGY, becomes marginally positive under both OpenDDE tracks.
Pepti-drift shows the most stable point estimates across the three pipelines
($-0.118$, $-0.114$, $-0.104$), whereas High-PepBinder and the
uniform-random control show the strongest negative associations throughout.
Because only point estimates are reported at this resolution, entries close to
zero should not be interpreted as method-level significance tests.

\begin{table}[htb]
\centering
\setlength{\belowcaptionskip}{4pt}
\caption{
Method-level within-target binding--safety association under three binding
oracles. The safety utility is the minimum signed margin over the four
liabilities defining the \safeactive{} endpoint. Entries are Spearman
$\rho$ point estimates over $n=2{,}416$--$2{,}640$ candidates per method and
oracle across 88 targets. Unlike Table~\ref{tab:oracle_robustness}, they carry
no confidence intervals and are computed without the benchmark's binary
structure-confidence and binding gates.
}
\label{tab:oracle_method_rho}
\small
\setlength{\tabcolsep}{5pt}
\begin{tabular}{@{}lrrr@{}}
\toprule
Method & ESMFold2+PRODIGY & OpenDDE+EvoEF2 & OpenDDE+HADDOCK3 \\
\midrule
Pepti-drift (ours)       & $-0.118$ & $-0.114$ & $-0.104$ \\
PepMLM                   & $-0.057$ & $-0.006$ & $-0.020$ \\
PepTune                  & $-0.084$ & $0.034$  & $0.034$  \\
MOG-DFM                  & $-0.081$ & $-0.062$ & $-0.090$ \\
High-PepBinder           & $-0.411$ & $-0.260$ & $-0.361$ \\
\midrule
Random uniform-length    & $-0.405$ & $-0.191$ & $-0.300$ \\
\bottomrule
\end{tabular}
\end{table}

The three binding oracles disagree much more strongly about individual
candidates than the aggregate statistics might suggest. Within
method--target strata, the Spearman correlation between candidate binding
scores is only $0.136$ for ESMFold2+PRODIGY versus OpenDDE+EvoEF2 and $0.207$
for ESMFold2+PRODIGY versus OpenDDE+HADDOCK3, rising to $0.569$ between the
two OpenDDE tracks that share a pose. Oracle disagreement is therefore
concentrated in \emph{which} candidate ranks highest rather than in the
aggregate direction of the binding--safety association.

We consequently interpret the repeated negative direction as robust to the
binding pipeline, while making no claim that the binding oracles are
interchangeable or that any predicted ranking establishes experimental
affinity. This establishes robustness to substantial changes on the binding
side of the evaluation; we next perform the complementary audit on the safety
side.

\subsection{Robustness to independent safety oracles}
\label{app:safety_oracles}

The preceding analysis perturbs the binding axis while holding safety fixed.
We now perform the complementary test: the primary ESMFold2+PRODIGY binding
score is held fixed while the safety utility is reconstructed from independent
predictors. We rescore toxicity and hemolysis using PeptiVerse's released
basic-task peptide-SMILES ChemBERTa CNN and sequence-based XGBoost models,
respectively~\citep{zhang2026peptiverse}, and obtain a second independent
hemolysis score from HemoPI2 Hybrid2 (ESM2-t6+MERCI, model 4)
~\citep{rathore2025prediction}. All external scores are complete for the 6,568 candidates
used in the primary binding--safety analysis, so every safety definition is
evaluated on exactly the same method--target--candidate panel. As in
Section~\ref{app:tradeoff}, binding and safety are ranked separately within
each method--target stratum before pooling.

Because the safety predictors use different released decision thresholds and
score calibrations, their raw outputs are not directly comparable. We therefore
define a threshold-normalized signed margin
\begin{equation}
\widetilde m(s;\tau)=
\begin{cases}
(\tau-s)/\tau, & s\leq\tau,\\
(\tau-s)/(1-\tau), & s>\tau,
\end{cases}
\label{eq:normalized_safety_margin}
\end{equation}
which maps the decision boundary to zero and the safe and unsafe score
endpoints to $+1$ and $-1$, respectively; larger values therefore indicate more
favorable predicted safety. We use $\tau=0.5$ for each internal head, 0.49 and
0.2801 for PeptiVerse toxicity and hemolysis, and 0.55 for HemoPI2 Hybrid2.

For each safety definition, the utility is the minimum of its included margins,
preserving the benchmark principle that a single adverse liability is
sufficient to make a candidate unsafe. \emph{PeptiVerse substitution} replaces
the internal toxicity and hemolysis margins with their PeptiVerse counterparts,
whereas \emph{HemoPI2 substitution} replaces only internal hemolysis.
\emph{External consensus} combines PeptiVerse toxicity and both external
hemolysis margins with the unchanged internal allergenicity and neurotoxicity
margins. Finally, \emph{all-oracle consensus} takes the minimum over all four
internal and all three external margins.

\begin{table}[htb]
\centering
\setlength{\belowcaptionskip}{4pt}
\caption{
Binding--safety association under five safety-oracle definitions with the
primary binding score held fixed. Safety is represented by the
threshold-normalized minimum margin in
Equation~\ref{eq:normalized_safety_margin}. The stratified estimate ranks
candidates within method--target strata before pooling; intervals are
2,000-sample target-bootstrap 95\% confidence intervals. The Internal row uses
exactly the C.1 candidate panel and recovers its stratified statistic.
}
\label{tab:safety_tradeoff_oracles}
\small
\setlength{\tabcolsep}{3.5pt}
\begin{tabular}{@{}lrrrr@{}}
\toprule
Safety definition &
$n$ &
Strat. $\rho$ &
\shortstack{Target-macro $\rho$\\{[95\% CI]}} &
\shortstack{Negative\\targets} \\
\midrule
Internal
& 6,568 & $-0.224$
& $-0.213$ [$-0.261$, $-0.162$] & 73/82 \\
PeptiVerse substitution
& 6,568 & $-0.094$
& $-0.091$ [$-0.141$, $-0.043$] & 56/82 \\
HemoPI2 substitution
& 6,568 & $-0.196$
& $-0.173$ [$-0.223$, $-0.125$] & 67/82 \\
External consensus
& 6,568 & $-0.096$
& $-0.093$ [$-0.139$, $-0.045$] & 55/82 \\
All-oracle consensus
& 6,568 & $-0.116$
& $-0.111$ [$-0.162$, $-0.061$] & 64/82 \\
\bottomrule
\end{tabular}
\end{table}

The central qualitative conclusion is preserved across all five safety
definitions: stronger predicted binding is associated with less favorable
predicted safety, and every target-macro confidence interval excludes zero.
HemoPI2 substitution remains closest to the internal estimate, whereas
PeptiVerse substantially attenuates the association without reversing its
direction. This result is not induced by the normalization in
Equation~\ref{eq:normalized_safety_margin}: using the corresponding
unnormalized signed margins yields stratified correlations of $-0.224$,
$-0.085$, $-0.202$, $-0.089$, and $-0.111$ in the same row order, again with
all target-macro confidence intervals excluding zero. Thus, the estimated
magnitude is safety-oracle dependent, but the negative direction is robust to
the independent oracle substitutions considered here.

Table~\ref{tab:safety_tradeoff_methods} resolves the same analysis by method
for the set of methods shown in the binding-oracle comparison.
Pepti-drift preserves a negative within-target association under all five
safety definitions ($\rho=-0.132$ to $-0.086$). High-PepBinder and the
uniform-random control are likewise consistently negative, whereas estimates
close to zero for PepMLM, PepTune, and MOG-DFM can change sign under
PeptiVerse-based definitions. These small method-level sign changes do not
alter the aggregate target-level conclusion and should not be interpreted as
individual significance tests.

\begin{table}[htb]
\centering
\setlength{\belowcaptionskip}{4pt}
\caption{
Method-level within-target binding--safety association as the safety oracle is
varied. Entries are Spearman $\rho$ point estimates using the
threshold-normalized safety margin over $n=545$--$929$ complete candidates per
method; no method-level confidence intervals are shown. The displayed methods
match Table~\ref{tab:oracle_method_rho} to facilitate comparison of sensitivity
to the binding and safety sides of the evaluation.
}
\label{tab:safety_tradeoff_methods}
\small
\setlength{\tabcolsep}{3.5pt}
\begin{tabular}{@{}lrrrrr@{}}
\toprule
Method & Internal & PV sub. & H2 sub. & Ext.\ cons. & All cons. \\
\midrule
Pepti-drift (ours)
& $-0.132$ & $-0.086$ & $-0.113$ & $-0.093$ & $-0.091$ \\
PepMLM
& $-0.015$ & $0.011$ & $-0.044$ & $-0.017$ & $-0.004$ \\
PepTune
& $-0.091$ & $-0.005$ & $-0.076$ & $-0.014$ & $-0.011$ \\
MOG-DFM
& $-0.110$ & $0.031$ & $-0.094$ & $0.058$ & $0.030$ \\
High-PepBinder
& $-0.400$ & $-0.212$ & $-0.326$ & $-0.204$ & $-0.244$ \\
\midrule
Random uniform-length
& $-0.444$ & $-0.248$ & $-0.416$ & $-0.256$ & $-0.332$ \\
\bottomrule
\end{tabular}
\end{table}

This directional robustness is nontrivial because the safety predictors are far
from interchangeable at the candidate level. Within method--target strata,
internal and PeptiVerse toxicity scores correlate at $-0.158$, internal and
PeptiVerse hemolysis scores at $-0.036$, and internal and HemoPI2 hemolysis
scores at $0.060$. The repeated negative binding--safety association therefore
cannot be explained by the external predictors simply reproducing the internal
scores. Instead, substantially different safety models disagree about which
individual candidates are safest while preserving the same aggregate
relationship: stronger predicted binders tend to occupy less favorable regions
of predicted safety.

Together with the binding-oracle analysis above, these results show that the
direction of the observed binding--safety association survives perturbations on
both sides of the computational evaluation pipeline. Its magnitude and the
ranking of individual candidates remain oracle-sensitive, so we do not treat
any evaluator as biological ground truth. 
These oracle substitutions show that the unadjusted negative association is not specific to a single evaluator. However, Figure~\ref{fig:tradeoff} shows that the association is nearly eliminated after exact-length stratification under the primary pipeline. We therefore do not interpret the cross-oracle result as evidence for an intrinsic length-independent binding--safety trade-off.
We next show that the same is true of its length mediation.

\subsection{Length adjustment of the oracle-robustness analyses}
\label{app:oracle_length}

We next test whether the length dependence in
Figure~\ref{fig:tradeoff} is robust to evaluator choice.
For each binding- and safety-oracle configuration, we repeat the
exact-length analysis by ranking binding and safety within method--target
strata and additionally stratifying by exact peptide length, requiring at
least three candidates per stratum. Confidence intervals use 2,000
target-cluster bootstrap replicates.

\begin{table}[htb]
\centering
\setlength{\belowcaptionskip}{4pt}
\caption{
\textbf{Length adjustment across binding and safety oracles.}
$b$ denotes binding score, $u$ safety utility, and $L$ peptide length.
Safety is the minimum signed margin over the four liabilities defining the
\safeactive{} endpoint.
}
\label{tab:oracle_length_adjusted}
\small
\setlength{\tabcolsep}{4pt}
\begin{tabular}{@{}lrrrrc@{}}
\toprule
Evaluator &
$n$ &
\shortstack{Unadjusted\\$\rho$} &
$\rho(b,L)$ &
$\rho(u,L)$ &
\shortstack{Exact-length $\rho$\\{[95\% CI]}} \\
\midrule

\multicolumn{6}{@{}l}{\emph{(a) Binding oracle varied; internal safety fixed}}\\
ESMFold2+PRODIGY
& 23,993 & $-0.235$ & $+0.469$ & $-0.395$
& $-0.014$ [$-0.033$, $+0.005$] \\
OpenDDE+EvoEF2
& 15,596 & $-0.098$ & $+0.228$ & $-0.343$
& $-0.002$ [$-0.024$, $+0.019$] \\
OpenDDE+HADDOCK3
& 15,596 & $-0.137$ & $+0.305$ & $-0.343$
& $-0.005$ [$-0.023$, $+0.014$] \\

\midrule
\multicolumn{6}{@{}l}{\emph{(b) Safety oracle varied; ESMFold2+PRODIGY binding fixed}}\\
Internal
& 6,568 & $-0.224$ & $+0.472$ & $-0.401$
& $+0.010$ [$-0.030$, $+0.048$] \\
PeptiVerse substitution
& 6,568 & $-0.094$ & $+0.472$ & $-0.166$
& $+0.014$ [$-0.031$, $+0.058$] \\
HemoPI2 substitution
& 6,568 & $-0.196$ & $+0.472$ & $-0.363$
& $-0.008$ [$-0.045$, $+0.029$] \\
External consensus
& 6,568 & $-0.096$ & $+0.472$ & $-0.174$
& $+0.022$ [$-0.025$, $+0.067$] \\
All-oracle consensus
& 6,568 & $-0.116$ & $+0.472$ & $-0.207$
& $+0.032$ [$-0.011$, $+0.073$] \\
\bottomrule
\end{tabular}
\end{table}

Exact-length stratification eliminates the association in all eight
configurations: adjusted correlations range from $-0.014$ to $+0.032$,
and every 95\% confidence interval contains zero.
The same conclusion holds with unnormalized safety margins, a minimum
stratum size of two, and each safety liability analyzed separately
(maximum adjusted $|\rho|=0.060$).

The magnitude of the unadjusted association also tracks each oracle's
dependence on peptide length. When safety is fixed, the three binding-oracle
estimates follow $\rho(b,L)$; when binding is fixed, the five safety-oracle
estimates follow $\rho(u,L)$. Thus, differences in the apparent
binding--safety association across evaluators are largely consistent with
differences in their length dependence.

Together, these results show that both the unadjusted negative association
and its removal by exact-length control are robust to evaluator choice.
We therefore interpret the apparent binding--safety trade-off as primarily
length-mediated rather than as evidence for a length-independent antagonism
between the two objectives.

\section{Additional Fixed-Budget Diagnostics}
\label{app:fixedbudget}

\subsection{Full fixed-budget joint results}
\label{sec:full_joint_results}

Table~\ref{tab:fixed_budget_joint} reports the numerical results underlying
the main-text quality--cost comparison. All methods are evaluated under the
same fixed budget of $K=30$ candidates per target over 88 held-out targets.
Raw \safeactive{} yield measures absolute joint success, whereas
Enrichment@30 reports the percentage-point improvement over the
target- and exact-length-matched random control. Target coverage is the
fraction of targets with at least one \safeactive{} candidate.
Generation throughput is measured under the protocol described in
Appendix~\ref{app:speed}.

\begin{table}[t]
\setlength{\belowcaptionskip}{4pt}
\caption{
\textbf{Full fixed-budget joint evaluation at $K=30$.}
SA denotes the percentage of all allocated candidate slots satisfying the
complete \safeactive{} endpoint. Enrichment is the percentage-point
improvement over the target- and exact-length-matched random control;
brackets show target-bootstrap 95\% confidence intervals. SA target coverage
is the percentage of held-out targets with at least one \safeactive{} hit.
}
\label{tab:fixed_budget_joint}
\centering
\small
\setlength{\tabcolsep}{4.5pt}

\begin{tabular}{lrrrr}
\toprule
Method &
SA $\uparrow$ &
Enrich. $\uparrow$ [95\% CI] &
SA target cov. $\uparrow$ &
peptides/sec $\uparrow$ \\
\midrule

\textbf{Pepti-drift (ours)}
& \underline{18.37}
& $\underline{+2.65\ [0.08,\,5.42]}$
& 60.23
& \textbf{378.3} \\

PepTune
& \textbf{18.45}
& $-0.49\ [-2.46,\,1.52]$
& 67.05
& 0.81 \\

MOG-DFM
& 10.98
& $-0.53\ [-2.50,\,1.40]$
& \underline{71.59}
& 0.01 \\

High-PepBinder
& 10.11
& $-2.05\ [-3.22,\,-0.83]$
& 65.91
& 0.73 \\

PepMLM
& 8.83
& $\mathbf{+5.11\ [1.97,\,8.67]}$
& 51.14
& 82.23 \\

AR Transformer
& 14.55
& $+0.83\ [-0.68,\,2.35]$
& \textbf{72.73}
& \underline{340.2} \\

\bottomrule
\end{tabular}
\end{table}

\subsection{Marginal safety and binding results}

The main comparison focuses on whether the same generated candidate satisfies
both binding and safety constraints. To clarify how this joint performance
arises, Tables~\ref{tab:safety_axis} and~\ref{tab:binding_axis} report the
corresponding marginal safety and binding results. These quantities are
descriptive rather than controlled leaderboards: the primary comparison remains
the fixed-budget \safeactive{} endpoint together with exact-length-matched
Enrichment@30.

\begin{table}[htb]
\setlength{\belowcaptionskip}{4pt}
\caption{
Predicted-safety results at $K=30$. All entries are percentages.
Safe-only is the fraction of submitted candidates satisfying sequence-integrity
requirements and all four safety gates without applying the binding criterion.
Target safe is the fraction of targets with at least one such candidate.
Tox, Hemo, All., and Neuro denote the fractions passing the toxicity,
hemolysis, allergenicity, and neurotoxicity gates, respectively.
Bold marks the best value and underline marks the second-best value among the
six noncollapsed learned or guided methods.
}
\label{tab:safety_axis}
\centering
\small
\setlength{\tabcolsep}{4.0pt}
\begin{tabular}{lrrrrrr}
\toprule
Method &
Safe-only &
Target safe &
Tox &
Hemo &
All. &
Neuro \\
\midrule
Pepti-drift (ours)
& \underline{61.17} & \underline{98.86}
& \underline{90.61} & \underline{94.55}
& 80.64 & \underline{82.23} \\

PepMLM
& 34.09 & 88.64 & 72.20 & 88.41 & 69.39 & 62.77 \\

PepTune
& \textbf{72.73} & \textbf{100.00}
& \textbf{96.86} & \textbf{96.14}
& \textbf{88.75} & \textbf{90.83} \\

AR Transformer
& 51.93 & \textbf{100.00} & 84.13 & 91.48 & 78.18 & 75.49 \\

MOG-DFM
& 57.90 & \textbf{100.00} & 88.76 & 90.02
& \underline{88.42} & 82.04 \\

High-PepBinder
& 42.88 & \textbf{100.00} & 78.41 & 88.18 & 78.52 & 64.62 \\

\midrule
Random length-matched
& 45.23 & 100.00 & 82.31 & 77.08 & 84.24 & 68.22 \\

Random uniform-length
& 35.68 & 100.00 & 71.33 & 70.11 & 84.85 & 57.39 \\

Train prior
& 0.00 & 0.00 & 85.83 & 92.65 & 78.41 & 78.11 \\
\bottomrule
\end{tabular}
\end{table}

\begin{table}[t]
\setlength{\belowcaptionskip}{4pt}
\caption{
Predicted-binding results at $K=30$.
Bind cov. is the percentage of submitted candidates receiving
a binding score after structure-confidence gating.
Bind-active additionally requires $\bindscore\geq7$;
both use all submitted candidates as the denominator
(2,640 per method; 2,634 for MOG-DFM).
Target pass thresholds each target's mean ipTM,
interface pLDDT, and $-\Delta G$ at 0.5, 70, and 7, respectively.
These means use all complete measurement triples before confidence
gating; all 88 targets are evaluable.
This diagnostic is distinct from any-hit target coverage.
Mean scored averages available confidence-gated binding scores;
ipTM and pLDDT average the corresponding target means.
Bold and underline indicate the best and second-best values
among the six noncollapsed learned or guided methods.
}
\label{tab:binding_axis}
\centering
\small
\setlength{\tabcolsep}{4.0pt}
\begin{tabular}{lrrrrrr}
\toprule
Method &
Bind cov. &
Bind-active &
Target pass &
Mean scored &
ipTM &
pLDDT \\
\midrule
Pepti-drift (ours)
& \textbf{35.45} & \underline{25.80} & \textbf{26.14}
& 8.20 & \textbf{0.442} & \underline{84.44} \\

PepMLM
& 31.10 & \textbf{27.16} & \underline{25.00}
& \textbf{9.51} & \underline{0.425} & 82.79 \\

PepTune
& \underline{32.54} & 24.32 & \underline{25.00}
& 8.39 & \underline{0.425} & \textbf{85.72} \\

AR Transformer
& 29.13 & 23.37 & 25.00
& 8.58 & 0.416 & 83.86 \\

MOG-DFM
& 23.46 & 18.75 & 15.91
& 8.68 & 0.372 & 83.35 \\

High-PepBinder
& 25.15 & 19.36 & 17.05
& \underline{8.82} & 0.386 & 83.27 \\

\midrule
Random length-matched
& 26.33 & 21.82 & 18.18 & 9.05 & 0.398 & 83.78 \\

Random uniform-length
& 20.64 & 16.74 & 9.09 & 9.17 & 0.370 & 82.51 \\

Train prior
& 25.80 & 20.15 & 19.32 & 8.79 & 0.391 & 83.64 \\
\bottomrule
\end{tabular}
\end{table}

\subsection{High-depth target analysis}
\label{app:discovery_depth}

The main-text analysis summarizes the distribution of \safeactive{} discovery
depth among reached targets. We further examine performance at increasingly
stringent depth thresholds by defining
\[
N_{\geq n}
=
\sum_{t\in\mathcal{T}}
\mathbf{1}[D_t \geq n],
\]
the number of held-out targets for which at least $n$ of the $K=30$ submitted
candidates satisfy the complete \safeactive{} endpoint.

\begin{figure}[t]
\centering
\includegraphics[width=0.70\textwidth]
{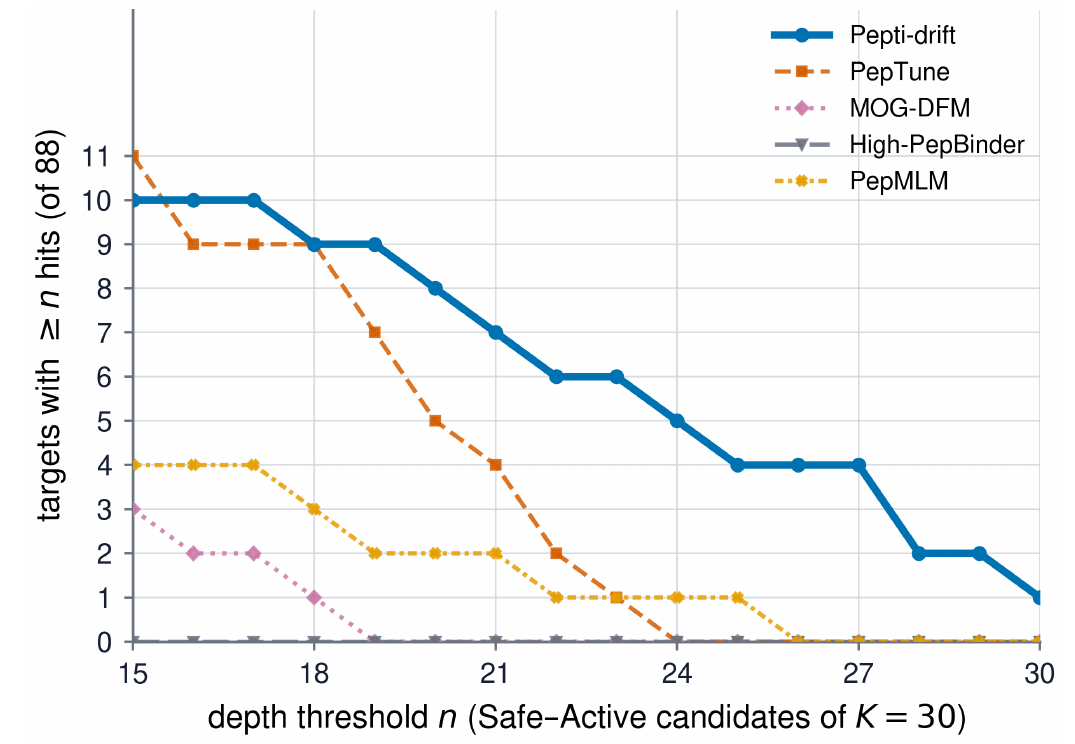}
\caption{
\textbf{Target-level discovery at high \safeactive{} candidate counts.}
For each threshold $n$, the plot reports the number of the 88 held-out targets
for which at least $n$ of $K=30$ submitted candidates satisfy the complete
\safeactive{} endpoint. Pepti-drift separates from the other methods at
high candidate-count thresholds and is the only method with any target
containing at least 26 \safeactive{} candidates.
}
\label{fig:depth_tail}
\end{figure}

Figure~\ref{fig:depth_tail} shows $N_{\geq n}$ for depth thresholds
$n=15,\ldots,30$. PepTune initially reaches one more target at $n\geq15$
(11 versus 10), and the two methods are tied at $n\geq18$ with nine targets
each. At more stringent thresholds, however, the separation increasingly
favors Pepti-drift. At $n\geq21$, Pepti-drift retains seven targets, compared
with four for PepTune and two for PepMLM. At $n\geq24$, Pepti-drift retains
five targets, whereas PepTune retains none and PepMLM retains one.

At the highest depth thresholds, Pepti-drift is the only method with any target
at $n\geq26$, retains four targets at $n\geq27$, and reaches $30/30$
\safeactive{} candidates on one target. Thus, the larger conditional mean
observed in the main text reflects consistently deeper feasible candidate pools
across multiple targets rather than being driven by a single exceptional
target.

\subsection{Mechanically selected target-level examples}
\label{app:target_examples}

To complement the aggregate discovery-depth analysis, we provide three
target-level examples selected by prespecified rules from the frozen
\safeactive{} counts. These examples are intended to illustrate distinct
target-level behaviors rather than to define the aggregate comparison.

For \emph{rescue} (rule R), we select a target reached by Pepti-drift but by
none of the other four published peptide generators considered here.
This rule identifies 2b1n\_A, on which Pepti-drift produces 2/30
\safeactive{} candidates while all four comparators produce none.

For \emph{depth} (rule D), we restrict attention to the 26 targets reached by
all five methods and rank them by Pepti-drift's depth advantage,
defined as its \safeactive{} count minus that of the strongest comparator
on the same target. The two largest gaps occur on 3cvl\_A and 2qa9\_E,
where the advantages are $+12$ and $+11$ candidates, respectively.
On 3cvl\_A, all 30 Pepti-drift submissions satisfy the complete joint
endpoint, compared with 18/30 for the strongest comparator; on 2qa9\_E,
the corresponding counts are 23/30 and 12/30.

\begin{figure*}[hbt]
\centering
\includegraphics[width=0.80\textwidth]
{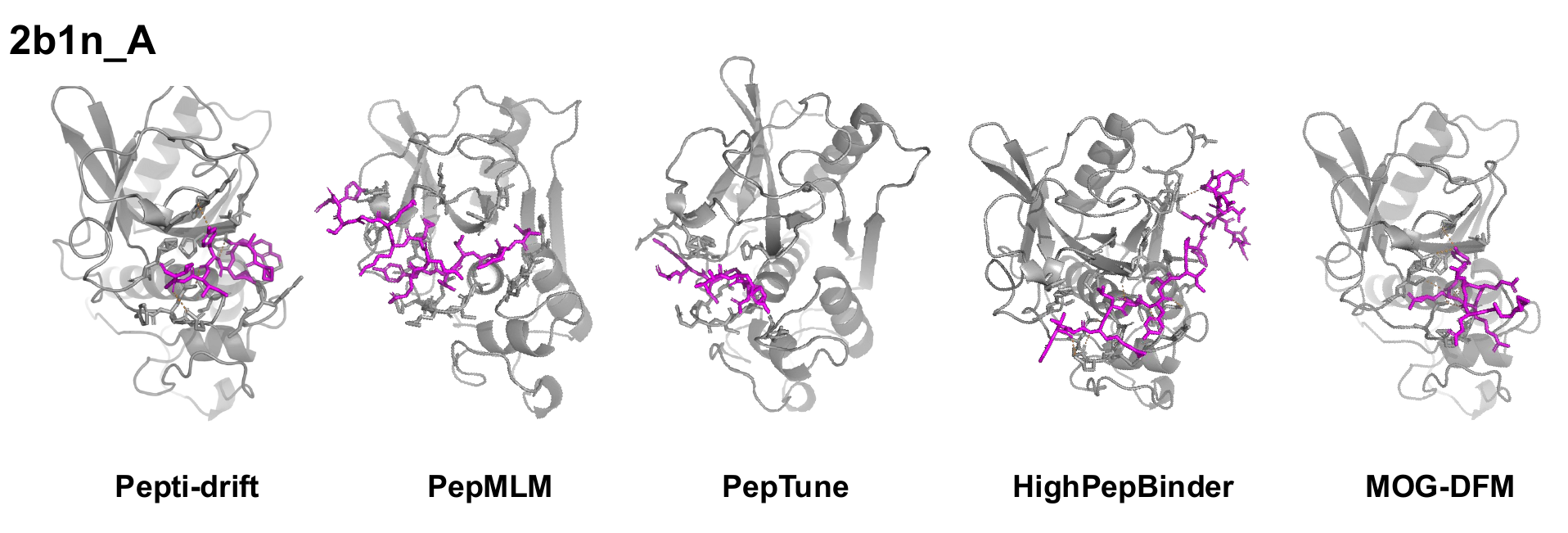}

\vspace{0.5em}

\scriptsize
\setlength{\tabcolsep}{4pt}
\renewcommand{\arraystretch}{1.08}

\resizebox{\linewidth}{!}{%
\begin{tabular}{llccccc}
\toprule
Method &
Peptide &
$\Delta G$ $\downarrow$ &
ipTM $\uparrow$ &
Toxicity $\downarrow$ &
Hemolysis $\downarrow$ &
\shortstack{Safe--Active\\hits/submitted} \\
\midrule

\multicolumn{7}{l}{\textit{Target 2b1n\_A} \; (rule R)} \\
\midrule
\textbf{Pepti-drift (ours)}
& {\ttfamily\scriptsize SAYNYIAV}
& $-10.5$ & $0.518$ & $0.001$ & $0.019$ & \textbf{2/30} \\

MOG-DFM
& {\ttfamily\scriptsize REEERRY}
& $-8.1$ & $0.448$ & $\mathbf{0.000}$ & $\mathbf{0.002}$ & $0/29$ \\

PepTune
& {\ttfamily\scriptsize VYNGCRS}
& $-8.9$ & $0.650$ & $0.011$ & $0.260$ & $0/30$ \\

High-PepBinder
& {\ttfamily\scriptsize GTPRLEYSIPHLNDDGETNN}
& $\mathbf{-12.5}$ & $0.682$ & $0.923$ & $0.013$ & $0/30$ \\

PepMLM
& {\ttfamily\scriptsize LKTTTYEENKQGSGH}
& $-11.4$ & $\mathbf{0.685}$ & $0.810$ & $0.012$ & $0/30$ \\

\midrule
\multicolumn{7}{l}{\textit{Target 3cvl\_A} \; (rule D, $+12$)} \\
\midrule

\textbf{Pepti-drift (ours)}
& {\ttfamily\scriptsize RFNPF}
& $-9.2$ & $0.938$ & $\mathbf{0.000}$ & $\mathbf{0.001}$ & \textbf{30/30} \\

MOG-DFM
& {\ttfamily\scriptsize YWKERIT}
& $-9.4$ & $0.874$ & $0.037$ & $0.052$ & $10/29$ \\

PepTune
& {\ttfamily\scriptsize VYFSKLD}
& $-9.6$ & $0.922$ & $0.007$ & $0.020$ & $18/30$ \\

High-PepBinder
& {\ttfamily\scriptsize LCSAID}
& $\mathbf{-10.3}$ & $\mathbf{0.941}$ & $0.000$ & $0.277$ & $9/30$ \\

PepMLM
& {\ttfamily\scriptsize QFSEELLLRLALKKL}
& $-9.5$ & $0.778$ & $0.264$ & $0.065$ & $4/30$ \\

\midrule
\multicolumn{7}{l}{\textit{Target 2qa9\_E} \; (rule D, $+11$)} \\
\midrule

\textbf{Pepti-drift (ours)}
& {\ttfamily\scriptsize ARPI}
& $-10.7$ & $\mathbf{0.958}$ & $\mathbf{0.000}$ & $\mathbf{0.119}$
& \textbf{23/30} \\

MOG-DFM
& {\ttfamily\scriptsize LLRYPLWRQTAWEGW}
& $\mathbf{-15.3}$ & $0.882$ & $0.071$ & $0.902$ & $10/30$ \\

PepTune
& {\ttfamily\scriptsize ISKLRDD}
& $-11.3$ & $0.933$ & $0.028$ & $0.224$ & $12/30$ \\

High-PepBinder
& {\ttfamily\scriptsize IHSLW}
& $-11.6$ & $0.939$ & $\mathbf{0.000}$ & $0.125$ & $9/30$ \\

PepMLM
& {\ttfamily\scriptsize DCVTYYIPPPAGLLK}
& $-13.6$ & $0.804$ & $0.357$ & $0.527$ & $1/30$ \\

\bottomrule
\end{tabular}%
}

\caption{
\textbf{Mechanically selected examples of target-level rescue and discovery
depth.}
Targets are selected from frozen per-target \safeactive{} counts using the
prespecified rules described in Appendix~\ref{app:target_examples}.
Rule R identifies 2b1n\_A, which is reached by Pepti-drift (2/30 hits) but
by none of the other four methods shown. Rule D selects the two largest
Pepti-drift depth advantages among the 26 targets reached by all five methods:
3cvl\_A ($30/30$ versus $18/30$ for the strongest comparator) and
2qa9\_E ($23/30$ versus $12/30$).
The structural examples above the table show target 2b1n\_A; within each
target--method pair, the displayed peptide is selected mechanically by maximum
finite ESMFold2 ipTM, with submitted rank breaking ties.
The final column reports \safeactive{} hits over the complete submitted set.
Bold indicates the best value within each target.
}
\label{fig:joint_discovery}
\end{figure*}

Figure~\ref{fig:joint_discovery} visualizes these mechanically selected
examples. Within each target--method pair, the displayed peptide is selected
independently of the \safeactive{} count by maximum finite ESMFold2 ipTM,
with submitted rank used only to break ties.

\begin{figure}[t]
\centering
\includegraphics[width=\textwidth]{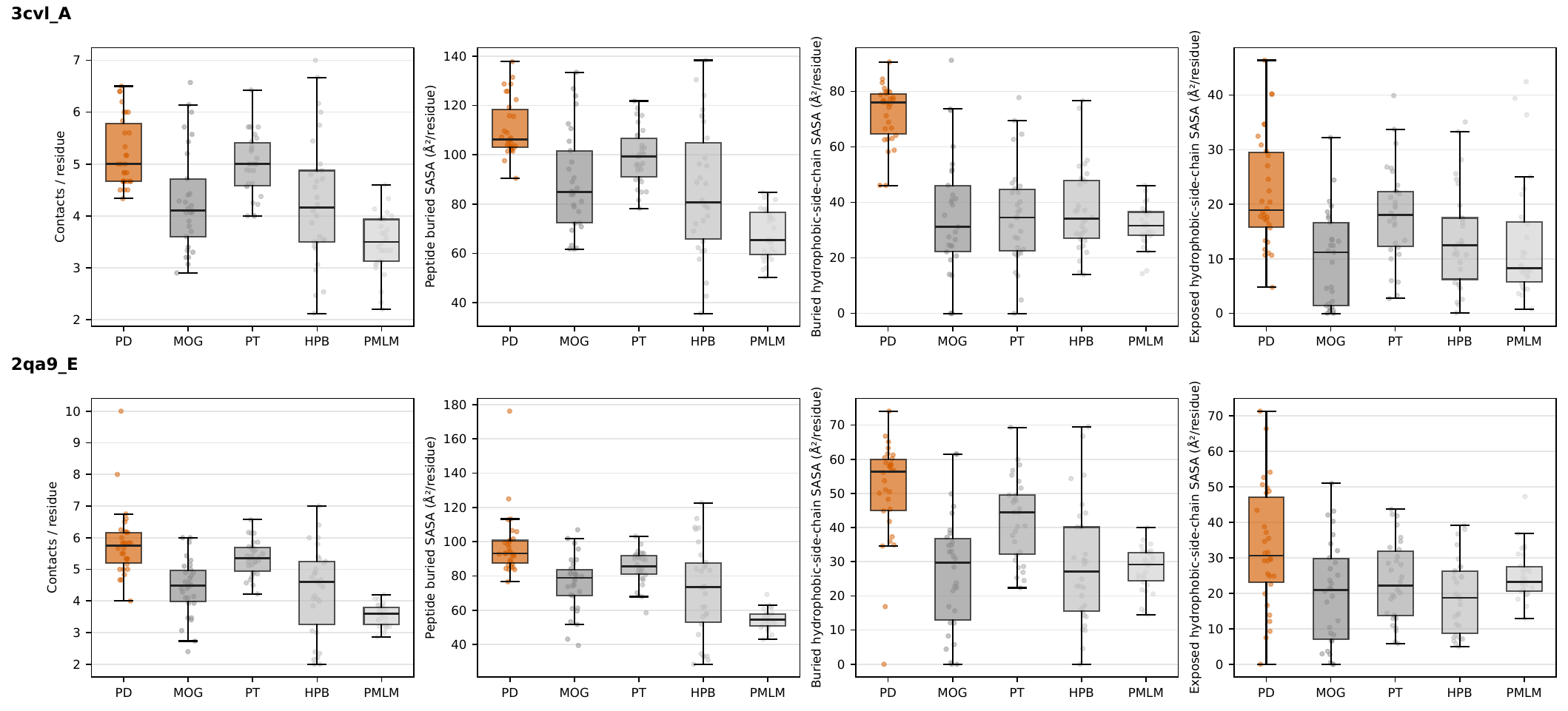}
\caption{\textbf{Targets with large Pepti-drift candidate pools exhibit stronger predicted interface utilization.}
Distributions across all submitted candidates are shown for 3cvl\_A and 2qa9\_E, the two targets with the largest Pepti-drift \safeactive{} candidate-count advantage among targets reached by all five methods.
Panels show contacts per residue, buried peptide SASA, buried hydrophobic-side-chain SASA, and exposed hydrophobic-side-chain SASA, with surface areas normalized per peptide residue.
Boxes show the interquartile range and median; points denote individual candidates.
PD, Pepti-drift; MOG, MOG-DFM; PT, PepTune; HPB, High-PepBinder; PMLM, PepMLM.}
\label{fig:contact_anal_pepti_drift}
\end{figure}

\begin{figure}[t]
\centering
\includegraphics[width=\textwidth]{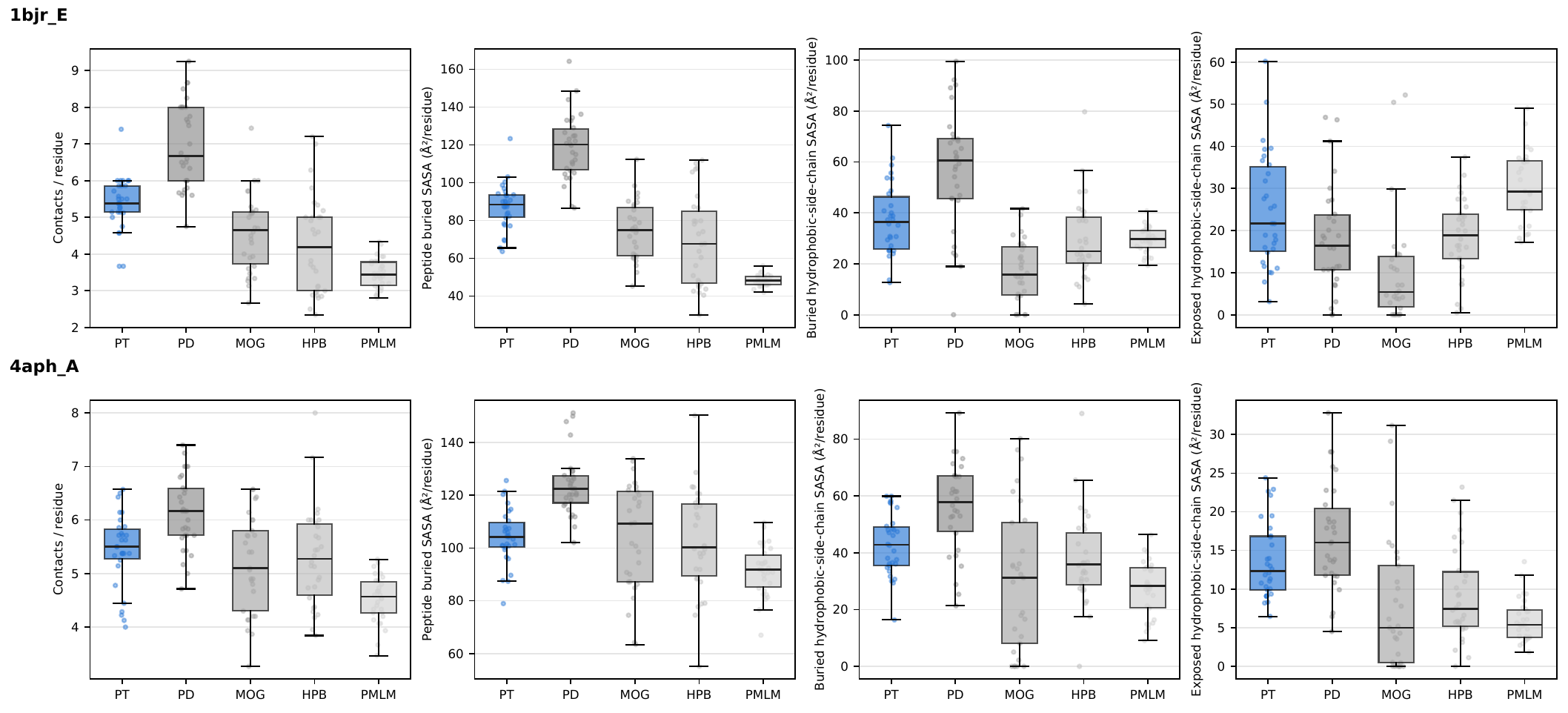}
\caption{
\textbf{Pepti-drift exhibits stronger predicted interface utilization on PepTune's top SAFE-ACTIVE targets.}
Distributions across all submitted candidates are shown for 1bjr\_E and 4aph\_A, the two targets with the largest \safeactive{} candidate counts for PepTune.
Panels show contacts per residue, buried peptide SASA, buried hydrophobic-side-chain SASA, and exposed hydrophobic-side-chain SASA, with surface areas normalized per peptide residue.
Boxes show the interquartile range and median; points denote individual candidates.
PT, PepTune; PD, Pepti-drift; MOG, MOG-DFM; HPB, High-PepBinder; PMLM, PepMLM.
}
\label{fig:contact_anal_peptune}
\end{figure}

\subsection{Predicted-structure interpretation of high-depth examples}
\label{app:structural_analysis}

We further examined the predicted structures of the two depth targets
mechanically selected in Appendix~\ref{app:target_examples}, 3cvl\_A and
2qa9\_E. Importantly, these targets were selected from the frozen
\safeactive{} counts before inspection of the structural descriptors.
Within each target--method pair, we use the same maximum-finite-ipTM
representative defined in Appendix~\ref{app:target_examples}. This analysis is
intended to interpret the benchmark predictions rather than to provide
independent validation of binding or biological safety.

\paragraph{Structural descriptors.}
A target--peptide residue contact was defined when any inter-chain heavy-atom
pair was within 4.5~\AA, with each residue pair counted once. The conclusions
were additionally checked at 4.0, 5.0, and 5.5~\AA. Solvent-accessible surface
area (SASA) was calculated with FreeSASA using ProtOr radii and a 1.4~\AA{}
probe. Peptide burial was defined as the SASA of the isolated peptide in its
bound conformation minus its SASA in the complex. We separately quantified
buried and exposed hydrophobic side-chain surface for AVILMFWYPC residues.
Because contact and surface-area quantities depend strongly on peptide length,
extensive quantities are reported per peptide residue.

\begin{table}[t]
\setlength{\belowcaptionskip}{4pt}
\caption{
\textbf{Structural descriptors of mechanically selected high-depth examples.}
Shown are the two targets selected by the prespecified discovery-depth rule,
3cvl\_A and 2qa9\_E. For each target--method pair, the representative peptide
is the fixed maximum-ipTM candidate. Contact density denotes the number of
target-residue contacts per peptide residue; buried SASA and exposed
hydrophobic side-chain SASA are normalized by peptide length; Hydro. SC burial
denotes the fraction of hydrophobic side-chain surface buried upon complex
formation. The final column reports the number of \safeactive{} candidates
among all submitted candidates for that target. These descriptors provide
structural interpretation of the predicted complexes rather than independent
binding validation.
}
\label{tab:structural_high_depth}
\centering

\setlength{\tabcolsep}{2.3pt}
\renewcommand{\arraystretch}{1.10}

\resizebox{\textwidth}{!}{
\begin{tabular}{lrrrrrrrrrr}
\toprule

& \multicolumn{5}{c}{\textbf{3cvl\_A}}
& \multicolumn{5}{c}{\textbf{2qa9\_E}} \\

\cmidrule(lr){2-6}
\cmidrule(lr){7-11}

Method &
\shortstack{Contact\\density $\uparrow$} &
\shortstack{Buried SASA\\(\AA$^2$/res) $\uparrow$} &
\shortstack{Hydro. SC\\burial $\uparrow$} &
\shortstack{Exposed hydro.\\SASA (\AA$^2$/res) $\downarrow$} &
\shortstack{Safe--Active\\hits/submitted $\uparrow$} &
\shortstack{Contact\\density $\uparrow$} &
\shortstack{Buried SASA\\(\AA$^2$/res) $\uparrow$} &
\shortstack{Hydro. SC\\burial $\uparrow$} &
\shortstack{Exposed hydro.\\SASA (\AA$^2$/res) $\downarrow$} &
\shortstack{Safe--Active\\hits/submitted $\uparrow$} \\

\midrule

\textbf{Pepti-drift (ours)}
& \textbf{6.40}
& \textbf{137.8}
& \textbf{79\%}
& \textbf{17.7}
& \textbf{30/30}
& \textbf{6.25}
& \textbf{112.8}
& \textbf{82\%}
& 13.8
& \textbf{23/30} \\

MOG-DFM
& 6.00
& 123.9
& 62\%
& 24.4
& 10/29
& 3.93
& 59.5
& 49\%
& 36.5
& 10/30 \\

PepTune
& 5.71
& 115.9
& 56\%
& 33.8
& 18/30
& 5.71
& 93.1
& 76\%
& \textbf{9.4}
& 12/30 \\

High-PepBinder
& 6.17
& 115.6
& 66\%
& 25.6
& 9/30
& 6.00
& 108.3
& 60\%
& 36.7
& 9/30 \\

PepMLM
& 2.53
& 56.9
& 54\%
& 25.0
& 4/30
& 4.07
& 62.2
& 47\%
& 36.8
& 1/30 \\

\bottomrule
\end{tabular}
}
\end{table}

\paragraph{3cvl\_A: productive hydrophobic burial.}
On 3cvl\_A, Pepti-drift's fixed representative RFNPF
(ipTM $=0.938$, predicted $\Delta G=-9.2$ kcal/mol, hemolysis score $=0.001$)
has the highest contact density among the five fixed representatives
(6.40 target-residue contacts per peptide residue) and the largest peptide
buried area (137.8~\AA$^2$ per residue). It buries 79.0\% of its hydrophobic
side-chain surface while leaving 17.7~\AA$^2$ per residue exposed.

This pattern is not restricted to the single displayed candidate.
Across all submitted Pepti-drift candidates on 3cvl\_A, the median contact
density is 5.00 per residue, the median buried peptide area is
106.2~\AA$^2$ per residue, and the median buried hydrophobic-side-chain SASA is
76.0~\AA$^2$ per residue. The latter is approximately 2.2-fold higher than
the largest competitor median. Relative to Pepti-drift's same-length random
controls, buried hydrophobic-side-chain SASA increases by
24.9~\AA$^2$ per residue on average, with 28 of 30 paired differences positive.
Thus, the high \safeactive{} depth on this target is accompanied by substantial
productive interfacial burial rather than being explained by peptide length
alone.

\paragraph{2qa9\_E: spatial allocation rather than lower sequence hydrophobicity.}
On 2qa9\_E, Pepti-drift's ARPI provides an illustrative contrast with the
MOG-DFM representative LLRYPLWRQTAWEGW. MOG-DFM has the stronger predicted
binding energy ($-15.3$ versus $-10.7$ kcal/mol) but a substantially less
favorable hemolysis score (0.902 versus 0.119). In the predicted complexes,
ARPI shows higher contact density (6.25 versus 3.93 contacts per residue),
greater buried peptide area (112.8 versus 59.5~\AA$^2$ per residue), and a
larger hydrophobic-side-chain burial fraction (81.6\% versus 48.8\%), while
leaving less hydrophobic-side-chain area exposed
(13.8 versus 36.5~\AA$^2$ per residue).

Notably, ARPI is not less hydrophobic at the sequence level: its GRAVY score
is 0.05 compared with $-0.63$ for the MOG-DFM representative, and its
hydrophobic-residue fraction is also higher (0.75 versus 0.60). This example
therefore supports an interpretation based on \emph{spatial allocation of
hydrophobicity}: hydrophobic residues can be retained but productively buried
at the predicted target interface rather than simply depleted from the
sequence.

\paragraph{Scope of interpretation.}
The reduced exposed hydrophobic surface observed for the fixed representatives
does not generalize uniformly across all submitted candidates. In particular,
Pepti-drift does not have the lowest median exposed hydrophobic-side-chain SASA
on both targets. We therefore interpret the consistent signal as strong
\emph{interface utilization} and \emph{productive hydrophobic burial}, rather
than a general tendency to minimize exposed hydrophobicity. Moreover, these
descriptors are calculated from the same ESMFold2 complexes used by the
benchmark binding pipeline and should therefore be viewed as qualitative
structural interpretation, not independent evidence of binding or a causal
mechanism for predicted safety.

\subsection{Integrity and omitted controls}

Pepti-drift generates 100\% canonical, in-range sequences with 100\%
within-target uniqueness, 99.51\% exact novelty relative to its generator
training peptides, and no exact cross-target reuse.

The integrity gates are important for several architecture controls.
Conditional discrete diffusion, the conditional VAE, and the
non-autoregressive Transformer have within-target uniqueness rates of 3.33\%,
1.33\%, and 0.15\%, respectively, and all three achieve zero \safeactive{}
yield and zero target coverage. The non-autoregressive Transformer emits only
four distinct sequences among 2,640 outputs despite nominally favorable values
on some marginal properties. The training-peptide prior likewise achieves zero
\safeactive{} yield because all candidates fail exact novelty.

The exact-length-matched random control reaches 10.27\% raw \safeactive{}
yield, 69.32\% target coverage, and 26.33\% confidence-gated binding coverage.
Its relatively strong raw yield further illustrates why raw \safeactive{} rate
alone is not sufficient for controlled comparison.

\subsection{Threshold sensitivity}

We re-gate the frozen $K=30$ candidate scores for all 12 methods and their
method-specific, same-length random controls.  No predictor or structure model
is rerun, and candidates with missing scores continue to fail the gate.  On the
joint $5\times5$ grid, the median Spearman
correlation with the default method ranking is 0.928 (minimum 0.840).  PepMLM
ranks first in 21 of 25 settings, whereas Pepti-drift retains positive
enrichment in 24 settings and remains in the top three throughout.

To identify which gate drives this variation, we additionally sweep one
threshold at a time while fixing the other four at their defaults.  The binding
minimum takes values $\{5,6,7,8,9\}$; each upper safety threshold takes values
$\{0.3,0.4,0.5,0.6,0.7\}$.  Unlike the joint sweep, this analysis separates the
effects of toxicity, hemolysis, allergenicity, and neurotoxicity.

\begin{figure}[t]
\centering
\includegraphics[width=\textwidth]{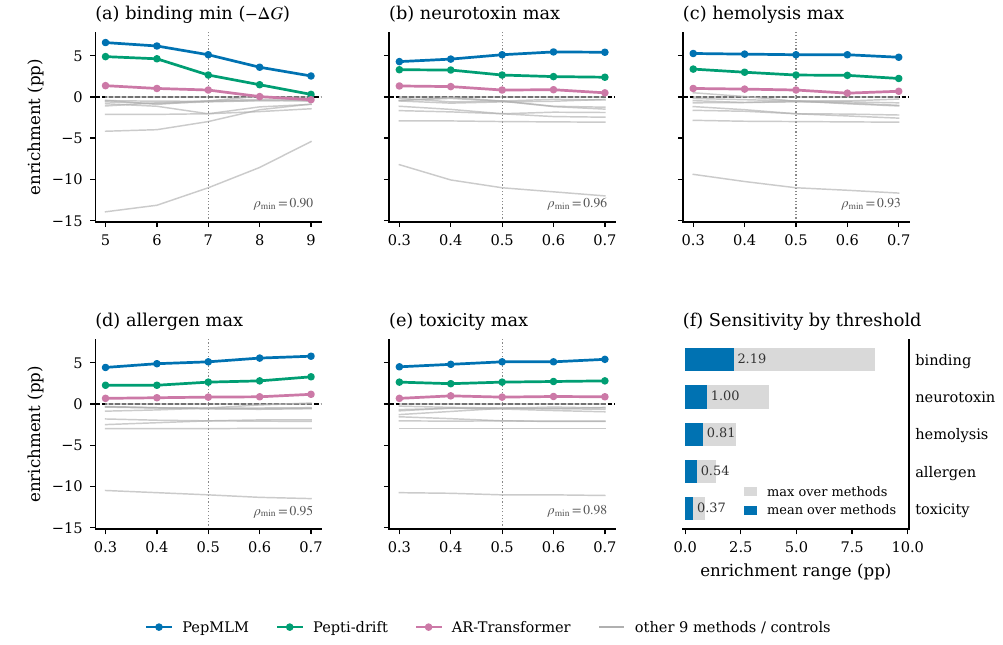}
\caption{\textbf{One-at-a-time sensitivity of Safe-Active Enrichment@$K$.}
Frozen scores are re-gated on 88 test targets while varying one threshold and
holding the other four at the benchmark default.  \textbf{(a)} Binding-score
minimum.  \textbf{(b--e)} Upper thresholds for neurotoxicity, hemolysis,
allergenicity, and toxicity.  Coloured curves show the three methods with
positive enrichment at the default gate; grey curves show the other nine
methods and controls.  Dotted lines mark the default threshold, and
$\rho_{\min}$ is the lowest Spearman correlation with the default 12-method
ranking in each sweep.  \textbf{(f)} The enrichment range induced by each
threshold, averaged over methods (blue) and for the most affected method
(grey).}
\label{fig:threshold-oat}
\end{figure}

Figure~\ref{fig:threshold-oat} shows that the binding cutoff is the dominant
source of threshold sensitivity.  Increasing the binding minimum from 5 to 9
reduces enrichment for each of the three leading methods; for Pepti-drift it
falls from $+4.89$ to $+0.34$ percentage points.  Nevertheless, the worst rank
correlation within this sweep remains high ($\rho_{\min}=0.900$).  Averaged
over methods, the binding sweep induces a 2.19 percentage-point enrichment
range, compared with 1.00 for neurotoxicity, 0.81 for hemolysis, 0.54 for
allergenicity, and 0.37 for toxicity.  Correspondingly, the individual safety
sweeps retain rank correlations of at least 0.928, with toxicity being nearly
inert ($\rho_{\min}=0.977$).

PepMLM remains the highest-enrichment method at every value in each
one-at-a-time sweep, and Pepti-drift remains positively enriched throughout.
The safety curves need not be monotone because enrichment is a paired
difference: relaxing a safety gate can admit additional candidates from both
the model and its same-length random control.  Together with the joint-grid
results, these sweeps show that the qualitative comparison is not determined
by one default cutoff.  They remain conditional on the frozen evaluators,
however, and cannot address evaluator misspecification or experimental
validity.

\section{Pepti-drift Architecture and Training Configuration}
\label{app:model_hyperparameters}

Table~\ref{tab:peptidrift_hparams} summarizes the architecture,
training, and optimization configuration of the main Pepti-drift checkpoint
used in the benchmark. The fixed peptide reference map follows the
construction described in Appendix~\ref{app:model_representation}.

\begin{table}[t]
\centering
\small
\setlength{\tabcolsep}{5pt}
\renewcommand{\arraystretch}{1.08}

\caption{
\textbf{Architecture and training hyperparameters for Pepti-drift.}
Values correspond to the main checkpoint used in the benchmark.
}
\label{tab:peptidrift_hparams}

\begin{tabular}{lll}
\toprule
Category & Hyperparameter & Value \\
\midrule

Representation
& ESM-2 encoder
& \texttt{esm2\_t36\_3B\_UR50D} \\
& ESM embedding dimension
& 2560 \\
& ESM pooling
& mean pooling \\
& ESM parameters
& frozen \\
& Target projection
& $2560\!\rightarrow\!2560\!\rightarrow\!2560$ \\
& Target projection activation
& ReLU \\
& Peptide reference projection
& $2560\!\rightarrow\!1024\!\rightarrow\!1024$ \\
& Peptide reference activation
& ReLU \\
& Peptide reference parameters
& fixed after random initialization \\
& Latent dimension $d_z$
& 1024 \\
& Latent normalization
& $\ell_2$ normalization \\
\midrule

Generator
& Architecture
& $3584\!\rightarrow\!2048\!\rightarrow\!2048\!\rightarrow\!1024$ \\
& Hidden activation
& ReLU \\
& Noise distribution
& $\mathcal{N}(0,I_{1024})$ \\
\midrule

Drift module
& Architecture
& $3584\!\rightarrow\!2048\!\rightarrow\!2048\!\rightarrow\!1024$ \\
& Hidden activation
& ReLU \\
& Number of drift steps
& 1 \\
& Drift step size $\eta$
& 1.0 \\
& Output gate
& learned scalar, initialized to 1.0 \\
\midrule

Decoder
& Architecture
& non-causal Transformer \\
& Maximum token slots $L$
& 50 \\
& Model dimension
& 1024 \\
& Transformer layers
& 4 \\
& Attention heads
& 8 \\
& Feed-forward dimension
& 4096 \\
& Transformer activation
& GELU \\
& Transformer normalization
& pre-norm \\
& Dropout
& 0.1 \\
& Positional representation
& learned queries \\
& Vocabulary
& 20 amino acids + special tokens \\
\midrule

Attraction--repulsion
& Attraction coefficient $\alpha_+$
& 1.0 \\
& Sampled negatives per example
& 64 \\
& Retained hard negatives $K_{\rm neg}$
& 32 \\
& Kernel temperature $\tau$
& 0.1 \\
& Final repulsion coefficient $\alpha_-$
& 1.0 \\
& Repulsion start $e_{\rm start}$
& 100 \\
& Repulsion warm-up $e_{\rm warmup}$
& 500 epochs \\
& Warm-up schedule
& linear \\
& Initial repulsion
& 0 \\
\midrule

Objective
& $\lambda_{\rm seq}$
& 1.0 \\
& $\lambda_{\rm drift}$
& 1024 \\
& $\lambda_{\rm reg}$
& 10.24 \\
\midrule

Optimization
& Optimizer
& AdamW \\
& Learning rate
& $1\times10^{-4}$ \\
& Weight decay
& $1\times10^{-4}$ \\
& Learning-rate scheduler
& none \\
& Batch size
& 32 \\
& Maximum epochs
& 1000 \\
& Early-stopping patience
& 10 epochs \\
& Early-stopping tolerance
& $1\times10^{-4}$ \\
& Gradient clipping
& none \\
& Random seed
& 42 \\
\bottomrule
\end{tabular}

\end{table}

\section{Reproducibility Details}
\label{app:reproducibility}

\subsection{Comparator submissions}

The eight comparators to Pepti-drift are PepMLM, PepTune, High-PepBinder,
MOG-DFM, conditional autoregressive and non-autoregressive Transformers, a
conditional VAE, and conditional discrete diffusion. The four bundled neural
baselines use mean-pooled ESM-2 3B target embeddings and the same strict binding
training export as Pepti-drift. We additionally evaluate a training-peptide
prior, random peptides sampled from the training length distribution, and
random peptides with length uniformly sampled from 4--25.

PepTune is restricted before submission to linear peptides containing the 20
standard L-amino acids. MOG-DFM generates ten candidates at each of length
quantiles 7, 10, and 15; six duplicated outputs leave 2,634 observed slots, and
the missing slots remain failures in fixed-budget metrics. High-PepBinder uses
its released checkpoint and preserves generation order. Unless a method
underfills its budget, each run contains 2,640 candidates.

The main leaderboard contains one declared generation run per method.
Accordingly, confidence intervals in the fixed-budget comparison quantify
variation across held-out targets rather than variation from repeated model
training.

\subsection{Binding-evaluator implementation}
\label{app:binding_evaluator}

For complex-structure prediction in BindSafe-PepBench, we use ESMFold2.
Specifically, we use the
\texttt{Synthyra/\allowbreak ESMFold2-\allowbreak Experimental-\allowbreak Fast-\allowbreak Cutoff2025} packaging of the
corresponding Biohub checkpoint. Interface binding energies are then computed
from the predicted complexes with PRODIGY (\texttt{prodigy-prot}
$\geq 2.4$)~\citep{xue2016prodigy}. All methods are evaluated with the same
checkpoint and inference configuration. Because the ESMFold2 repository is
an actively updated remote-code model tracking \texttt{main}, the checkpoint is
resolved once and reused for every method rather than refetched per run.

\subsection{Pretrained weights and external checkpoints}
\label{app:pretrained_weights}

Every pretrained checkpoint used anywhere in the benchmark is fixed for all
methods and listed below. Weights are resolved from the Hugging Face Hub unless
stated otherwise.

\begin{itemize}
\item \textbf{Complex-structure prediction (primary binding oracle).}
  \texttt{Synthyra/\allowbreak ESMFold2-\allowbreak Experimental-\allowbreak Fast-\allowbreak Cutoff2025}, as described above.
\item \textbf{Target encoder for Pepti-drift and the bundled baselines.}
  ESM-2 3B, \texttt{facebook/\allowbreak esm2\_\allowbreak t36\_\allowbreak 3B\_\allowbreak UR50D}, mean-pooled over the
  target sequence. The same embeddings are used by Pepti-drift, the conditional
  autoregressive and non-autoregressive Transformers, the conditional VAE, and
  conditional discrete diffusion; these five models carry no other pretrained
  weights and are trained in this work from the strict binding training export.
\item \textbf{Safety and property evaluators.} ESM-2 650M,
  \texttt{facebook/\allowbreak esm2\_\allowbreak t33\_\allowbreak 650M\_\allowbreak UR50D}, used as a frozen mean-pooling
  encoder under a lightweight MLP head trained in this work for each of the seven
  properties (toxicity, hemolysis, allergenicity, neurotoxicity, nonfouling,
  cell penetration, and hemolytic HC$_{50}$). The four properties defining the
  \safeactive{} gate reach held-out ROC-AUC of $0.756$ (toxicity), $0.825$
  (hemolysis), $0.874$ (allergenicity), and $0.700$ (neurotoxicity).
\item \textbf{PepMLM.} \texttt{TianlaiChen/\allowbreak PepMLM-650M}, the authors' released
  checkpoint.
\item \textbf{MOG-DFM.} The gated \texttt{ChatterjeeLab/\allowbreak MOG-DFM} release,
  comprising the peptide flow-matching model and the authors' guidance
  classifiers for binding affinity, hemolysis, nonfouling, and half-life. Its
  internal sequence encoder is ESM-2 650M, the same checkpoint as above.
\item \textbf{PepTune.} The authors' released masked discrete diffusion
  checkpoint together with their SMILES SPE tokenizer, PeptideCLM embedder, and
  their own property classifiers. PepTune runs in its own container and is the
  only comparator whose scores originate from a SMILES-level model.
\item \textbf{High-PepBinder.} The released latent-diffusion checkpoint
  \texttt{pepldm\_\allowbreak step=882-\allowbreak val\_\allowbreak loss=0.51.\allowbreak ckpt} (5.69\,GB, Zenodo record
  \texttt{18282817}, MD5 \texttt{e98b1fc1\allowbreak 2a0f0c73\allowbreak f5977519\allowbreak 6e9428f3}), used with
  its released batch size of 64.
\item \textbf{Independent binding oracles.} The OpenDDE~v1 checkpoint
  \texttt{opendde.pt} generates the poses shared by the OpenDDE+EvoEF2 and
  OpenDDE+HADDOCK3 tracks of Appendix~\ref{app:binding_oracles}; EvoEF2 and
  HADDOCK3 are scoring functions and carry no learned weights.
\end{itemize}

The training-peptide prior and the two random-peptide controls use no pretrained
weights.

\subsection{Generation-speed protocol}
\label{app:speed}
Generation time is measured separately for each of the 88 held-out targets on
one NVIDIA H100 80GB HBM3, from the raw target amino-acid sequence to the
decoded peptide strings. The timed region includes target preprocessing and
embedding required by the method, sampling or inference-time property guidance,
decoding, native validity filtering, and any additional sampling needed to
reach the requested budget. Model and tokenizer loading, checkpoint loading,
disk I/O, and one target-independent warm-up are excluded.

During leaderboard generation, Pepti-drift and the autoregressive Transformer
can read precomputed ESM-2 target embeddings. For the timing comparison, both
instead re-embed every target inside the timed region so that their target
encoding cost is treated consistently with the other sequence-conditioned
methods. Target embedding accounts for approximately 80\% of Pepti-drift's
measured runtime and 72\% of the autoregressive Transformer's runtime.

\begin{table}[t]
\centering
\setlength{\belowcaptionskip}{4pt}
\caption{
Generation speed for producing $K=30$ candidates per held-out target on one
NVIDIA H100. Per-candidate cost divides by the requested budget of 30 rather
than by the number returned, so under-delivery is not rewarded. Relative cost
is normalized to Pepti-drift.
}
\label{tab:generation_speed}
\centering
\small
\setlength{\tabcolsep}{3pt}
\begin{tabular*}{\linewidth}{@{\extracolsep{\fill}}lrrrr@{}}
\toprule
Method & Time/target (s) & Time/candidate (ms) &
Candidates/s & Relative \\
\midrule
MOG-DFM
& 2{,}492.0 & 83{,}065.8 & 0.012 & $31{,}426.8\times$ \\
High-PepBinder
& 40.85 & 1{,}361.5 & 0.734 & $515.1\times$ \\
PepTune
& 37.08 & 1{,}236.0 & 0.809 & $467.6\times$ \\
PepMLM
& 0.365 & 12.16 & 82.23 & $4.6\times$ \\
AR Transformer
& 0.088 & 2.940 & 340.2 & $1.1\times$ \\
Pepti-drift (ours)
& \textbf{0.079} & \textbf{2.643} & \textbf{378.3} & $1.0\times$ \\
\bottomrule
\end{tabular*}
\end{table}

Pepti-drift, PepMLM, the autoregressive Transformer, and High-PepBinder were
measured under the common timing harness. The three fastest methods were
averaged over five complete passes over the 88 targets. PepTune and MOG-DFM use
the per-target clocks recorded by their original $K=30$ generation runs under
the same model-loading and I/O exclusions. MOG-DFM was not rerun solely for
timing because doing so would require approximately 61 additional GPU-hours;
its original timer also used wall-clock time without an explicit CUDA
synchronization. These implementation differences should be considered when
interpreting precise runtime ratios, although they do not affect the
orders-of-magnitude separation from Pepti-drift.

The measured costs also reflect constraints required by the benchmark. PepTune
continues sampling until 30 parseable linear canonical peptides are obtained.
MOG-DFM assembles $K=30$ from three separate length-specific generation calls.
High-PepBinder uses its released batch size of 64 and therefore oversamples the
requested budget of 30.

\end{document}